\documentclass[a4paper,conference]{IEEEtran}
\usepackage[left=2.22cm,right=2.22cm,top=2.10cm,bottom=2.7cm]{geometry}

\pdfpageattr{/Group << /S /Transparency /I true /CS /DeviceRGB >>}

\usepackage[utf8]{inputenc}
\usepackage[T1]{fontenc}
\usepackage{textcomp}
\usepackage{microtype}
\usepackage{calc}
\usepackage[normalem]{ulem}
\usepackage{verbatim}

\usepackage{lipsum}

\usepackage[range-phrase=--,per-mode=symbol-or-fraction,range-units=single,list-units=single,detect-all,forbid-literal-units=true]{siunitx}
\AtBeginDocument{
\DeclareSIUnit\bit{bit}
\DeclareSIUnit\byte{Byte}
\DeclareSIUnit\mbps{\mega\bit\per\second}
\DeclareSIUnit\kmh{\kilo\meter\per\hour}
\DeclareSIUnit\mw{\milli\watt}
\DeclareSIUnit\decibelm{dBm}
\DeclareSIUnit\decibeli{dBi}
\DeclareSIUnit\vehicle{veh}
}

\usepackage{silence}
\usepackage{csquotes}
\usepackage[backend=biber,style=ieee,doi=false,isbn=false,mincitenames=1,maxcitenames=2]{biblatex}
\DeclareFieldFormat{sentencecase}{#1} 
\DeclareFieldFormat{titlecase}{#1} 
\usepackage{xpatch}
\xpatchbibmacro{textcite}{\addspace}{\addnbspace}{}{}
\xpatchbibmacro{Textcite}{\addspace}{\addnbspace}{}{}
\DefineBibliographyStrings{english}{
	andothers = et~al\adddot\addspace
}

\DeclareSourcemap{
	\maps[datatype=bibtex]{
		\map[overwrite=true]{
			\step[fieldsource=school,fieldtarget=institution]
		}
	}
}

\usepackage{amssymb,amsmath,amsthm}
\usepackage{amssymb}
\usepackage{amsfonts}
\usepackage{amsthm}
\usepackage{comment}
\usepackage{algorithm}
\usepackage{algpseudocode}

\usepackage{algpseudocode}

\usepackage{booktabs}
\usepackage{url}

\usepackage[inline]{enumitem}

\usepackage[caption=false,font=footnotesize]{subfig}
\usepackage[pdftex]{graphicx}
\DeclareGraphicsExtensions{.pdf,.png,.jpg}
\usepackage{tikz}
\usetikzlibrary{arrows}
\usetikzlibrary{calc}
\usetikzlibrary{chains}
\usetikzlibrary{scopes}

\usepackage[american]{babel}
\usepackage{hyphenat}

\usepackage[capitalize,noabbrev,nameinlink]{cleveref}

\RequirePackage{xstring}
\RequirePackage{xparse}
\RequirePackage[]{acro}
\makeatletter
\@ifpackagelater{acro}{2019/02/17}{
	\NewDocumentCommand\acrodef{mO{#1}mG{}}{\DeclareAcronym{#1}{short={#2}, long={#3}, foreign-plural={}, #4}}
}{
	\NewDocumentCommand\acrodef{mO{#1}mG{}}{\DeclareAcronym{#1}{short={#2}, long={#3}, #4}}
}
\makeatother

\acrodef{AI}{Artificial Intelligence}
\acrodef{BER}{Bit Error Rate}
\acrodef{CSI}{Channel State Information}
\acrodef{IoT}{Internet of Things}
\acrodef{LBT}{Listen-Before-Talk}
\acrodef{MCS}{Modulation and Coding Scheme}
\acrodef{NG-TCMS}{Next-Generation Train Control and Monitoring System}
\acrodef{NR}{New Radio}
\acrodef{QoS}{Quality of Service}
\acrodef{SCI}{Sidelink Control Information}
\acrodef{SCS}{Subcarrier Spacing}
\acrodef{SL}{Sidelink}
\acrodef{SNR}{Signal-to-Noise-Ratio}
\acrodef{SPS}{Semi-Persistent Scheduling}
\acrodef{RSRP}{Reference Signal Received Power}
\acrodef{TB}{Transport Block}
\acrodef{TSN}{Time-Sensitive Networking}
\acrodef{UE}{User Equipment}
\acrodef{V2I}{Vehicle-to-Infrastructure}
\acrodef{V2P}{Vehicle-to-Pedestrian}
\acrodef{V2V}{Vehicle-to-Vehicle}
\acrodef{V2X}{Vehicle-to-Everything}
\acrodef{WLCN}{WireLess Consist Network}
\acrodef{WLTB}{WireLess Train Backbone}
\acrodef{PRR}{Packet Reception Ratio}
\acrodef{PRB}{Physical Resource Block}
\acrodef{SINR}{Signal-to-Interference-plus-Noise-Ratio}
\acrodef{gNB}{g-NodeB}
\acrodef{LoS}{Line-of-Sight}

\acrodef{B.A.T.M.A.N.}{Better Approach To Mobile Ad-hoc Networking}
\acrodef{TTL}{Time to Live}
\acrodef{RB}{Resource Block}
\acrodef{OGM}{Originator Message}
\acrodef{MPR}{Multipoint Relaying}
\acrodef{PDR}{Packet delivery ratio}
\acrodef{D2D}{Device-to-Device}
\acrodef{OLSR}{Optimized Link State Routing Protocol}
\acrodef{AODV}{Ad hoc On-Demand Distance Vector Protocol}

\usepackage{todonotes}

\def\todoCtd#1{%
	TODO: #1%
	\ifx&#1&...\fi%
	\endgroup
	\relax
}

\NewDocumentCommand\IEEE{ s m >{\SplitArgument{4}{/}}d[] }{%
	\IfBooleanTF{#1}{}{IEEE\,}
	\nolinebreak[2]
	#2%
	\IfNoValueTF{#3}{%
	}{%
		\sommerIEEELettersSlashed#3%
	}%
}
\newcommand{\sommerIEEELettersSlashed}[5]{%
	\IfNoValueTF{#2}{%
	}{%
		\nolinebreak[3]
	}%
	#1%
	\IfNoValueTF{#2}{}{/#2}%
	\IfNoValueTF{#3}{}{/#3}%
	\IfNoValueTF{#4}{}{/#4}%
	\IfNoValueTF{#5}{}{/#5}%
}

\usepackage{pifont}

\newcommand{\supp}{\textit{supp}}
\newcommand{\contr}{\textit{contr}}

\newtheorem{theorem}{Theorem}
\newtheorem{lemma}{Lemma}
\newtheorem{proposition}{Proposition}
\newtheorem{definition}{Definition}

\newtheorem{policy}{Policy}

\begin{document}

\title{Belief Graphs with Reasoning Zones:\\ Structure, Dynamics, and Epistemic Activation}

\author{%
\IEEEauthorblockN{%
    Saleh Nikooroo and Thomas Engel
}%

\small{
    \texttt{%
	saleh.nikooroo%
	@uni.lu, thomas.engel@uni.lu%
    }
}
\\
}

\maketitle


\begin{abstract}
\noindent Belief systems are rarely globally consistent, yet effective reasoning often persists locally. We propose a novel graph-theoretic framework that cleanly separates \emph{credibility}—external, a priori trust in sources—from \emph{confidence}—an internal, emergent valuation induced by network structure. Beliefs are nodes in a directed, signed, weighted graph whose edges encode support and contradiction. Confidence is obtained by a contractive propagation process that mixes a stated prior with structure-aware influence and guarantees a unique, stable solution.
Within this dynamics, we define \emph{reasoning zones}: high-confidence, structurally balanced subgraphs on which classical inference is safe despite global contradictions. We provide a near-linear procedure that seeds zones by confidence, tests balance using a parity-based coloring, and applies a greedy, locality-preserving repair with Jaccard de-duplication to build a compact atlas.
To model belief change, we introduce \emph{shock updates} that locally downscale support and elevate targeted contradictions while preserving contractivity via a simple backtracking rule. Re-propagation yields localized reconfiguration—zones may shrink, split, or collapse—without destabilizing the entire graph.
We outline an empirical protocol on synthetic signed graphs with planted zones, reporting zone recovery, stability under shocks, and runtime. The result is a principled foundation for contradiction-tolerant reasoning that activates classical logic precisely where structure supports it.
\end{abstract}


\acresetall
\IEEEpeerreviewmaketitle

%


\section{Introduction}
\label{sec:introduction}

Belief systems, whether human, artificial, or hybrid, are rarely globally coherent. In practical reasoning, contradictions, incomplete information, and structural uncertainty are the rule rather than the exception. Yet agents still reason and decide effectively. This motivates a central question: {How can reasoning operate meaningfully when the underlying belief substrate is fragmented or partially inconsistent?}

Classical approaches (Bayesian models \cite{10131168}, \cite{Tianjin_Li_2023}, doxastic logics \cite{10.1093/oso/9780192898852.001.0001}, AGM-style revision \cite{Hansson2023}) have been foundational, but they typically presuppose globally consistent knowledge bases or compress belief into scalar degrees detached from internal structure. They struggle to model belief as a \emph{system} in which propositions interact through relations of support and contradiction and where uncertainty propagates along those relations.

We propose a graph-theoretic framework that separates \emph{credibility}—externally assigned, a priori trust in sources—from \emph{confidence}—an endogenous, structure-induced valuation of beliefs. Beliefs are nodes in a directed, signed, weighted graph which edges encode \emph{evidential influence}: positive edges (support) push beliefs upward and negative edges (contradiction) push them downward. Edge weights quantify influence strength; they are not logical entailments or probabilities of implication. Confidence values for nodes are obtained by a damped propagation process that (i) normalizes support and contradiction by outgoing mass, (ii) mixes a stated prior with graph influence, and (iii) is tuned to be contractive, yielding a unique, stable solution for every graph instance.  

A key concept is that of reasoning zones (RZs): confidence-thresholded, locally consistent subgraphs in which signed cycles are non-contradictory. Concretely, a candidate zone passes a Harary-style parity (balance) test {that can be checked in linear time via signed 2-coloring} \cite{10.1307/mmj/1028989917}. Within an RZ, classical inference can run safely even if the global graph remains inconsistent. Multiple RZs may coexist and overlap; we maintain an \emph{atlas} (compact set of high-quality zones) using a  governance policy that de-duplicates near-duplicates by  Jaccard overlap and prefers higher-confidence, denser subgraphs.

We also introduce a dynamic update model to handle incoming information, including contradictions. A \emph{shock rule} locally downscales outgoing support from affected beliefs and raises targeted contradictions, followed by re-propagation under the same contraction safeguards. This produces localized reconfiguration: zones may shrink, split, or collapse without global oscillation or loss of stability.

\subsection*{Contributions}
\begin{itemize}
    \item We introduce a formal belief graph with two layers: \emph{credibility} (external priors over nodes) and \emph{confidence} (structure-derived valuations). Confidence is computed by a normalized, damped propagation scheme with guaranteed convergence under a stated contractivity condition.
    \item We define \emph{reasoning zones} as confidence-stable, locally consistent (balanced) subgraphs, and propose a near-linear extraction pipeline: thresholding by confidence, linear-time balance testing via signed 2-coloring, greedy locality-preserving repair, and atlas assembly with explicit Jaccard de-duplication.
    \item We design a governance mechanism for overlapping zones (the zone atlas) and a safe contradiction \emph{shock} update with backtracking to preserve contractivity, enabling stable adaptation under new evidence.
    \item We present an empirical plan on synthetic signed graphs with planted zones, reporting zone recovery and stability under shocks, redundancy vs. coverage in the atlas, and convergence iterations/runtime.
    \item We outline a lightweight interface that passes each zone to a downstream reasoning module, enabling selective, contradiction-tolerant inference without requiring global consistency.
\end{itemize}

\noindent This framework marks an initial step toward contradiction-tolerant inference, selective reasoning, and epistemic robustness in dynamic, fragmented environments, while remaining compatible with standard reasoning architectures.

%
\section{Background and Related Work}
\label{sec:background}
  
The representation and manipulation of belief has been a central concern across logic, artificial intelligence, epistemology, and decision theory. Traditional approaches typically treat beliefs as either sets of propositions or numerical degrees of certainty, with reasoning assumed to operate over globally coherent systems. This section reviews  contemporary perspectives across logic-based, probabilistic, structural, and cognitive models of belief, with special attention to their representational assumptions, limitations, and relation to our proposed framework.

Recent research on belief representation and epistemic structure has seen a growing interest in graph-based models that go beyond traditional probabilistic or logical paradigms. Below, we discuss closely related efforts across structural belief modeling, uncertainty handling, and authority-aware reasoning.

\subsection{Graph-based models of belief systems}

Several recent approaches have modeled beliefs as nodes in a directed graph, emphasizing structural dynamics and interdependencies. The BENDING model \parencite{visceanu2023bending} presents a framework for understanding collective and individual belief resilience through cognitive graph structures. Similarly, \textcite{dumbrava2025semantic} proposes a modular cognitive architecture that frames belief as structured semantic states within interpretable AI agents. Though less focused on belief per se, models based on causal graph relationships echo our aim of structurally exposing epistemic dependency patterns. Our approach aligns with these in using directed, weighted graphs, but diverges in its focus on fine-grained epistemic tension, dual-source modeling, and static representational foundations rather than dynamic updating.
Our model builds on our preliminary formalizations of belief graphs reported in \parencite{nikooroo2025_belief_system}, and localized reasoning structures reported in \parencite{nikooroo2025_reasoning_system}, extending them with a unified treatment of structural coherence and epistemic dynamics.

\subsection{Reasoning over epistemic and knowledge graphs}

Graph-based reasoning has also been explored in formal and computational epistemology. \textcite{hunter2022epistemic} introduces a SAT-based framework for reasoning over epistemic graphs, using constraints on belief propagation in argumentative structures. In contrast, our framework does not impose logical closure and avoids constraint solving, instead emphasizing the representational expressivity of local epistemic support and conflict. Other works, such as \textcite{jacquin2023uncertainty}, use belief function theory to model epistemic uncertainty over knowledge graphs, highlighting the need for flexible structures in uncertain settings—an objective shared with our confidence–credibility decomposition.

\subsection{Modeling external trust and authority}

A distinctive feature of our model is the explicit separation between internal confidence and external credibility. This mirrors ideas in authority-weighted epistemology such as the BEWA framework \parencite{wright2025bewa}, which uses graph-based propagation and cryptographic arguments to enforce trustworthy reasoning in autonomous agents. Similarly, \textcite{hunter2022trust} develops trust graphs for belief revision using ordinal conditional functions, demonstrating the utility of relational trust modeling in belief systems. While both works target belief dynamics and revision, our formulation provides a minimal, static representation layer upon which such mechanisms could be built. Jiang~\cite{jiang2024trustbased} offers a logic of trust-based beliefs that bypasses global consistency by grounding belief in data trust relationships. Chen~\cite{chen2022statemental} introduces a credibility logic framework based on algebraic reasoning under uncertainty, generalizing classical Bayesian inference with robust semantic inference calculus. Hunter~\cite{hunter2023trust} expands on this line by developing a dual-mechanism trust-update system using agent performance and AGM-style revision.

\subsection{Topological and probabilistic extensions}

Alternative frameworks have extended belief modeling using topological or probabilistic approaches. Prieto~\cite{prieto2024dempstershafer} proposes a belief model combining Dempster-Shafer theory with topological evidence structures. Shenoy~\cite{shenoy2023distinct} examines the structure of distinct belief functions within graphical models, and Delgrande~\cite{delgrande2023epistemic} presents a logic for epistemic likelihoods where beliefs carry sub-certainty weight. Though conceptually rich, such models often emphasize scalar representation of belief strength, whereas our model preserves structural heterogeneity without requiring numerical fusion.

\subsection{Argument Structures and Inference under Uncertainty}

Several recent approaches have explored inference and belief management under complex or uncertain knowledge graph structures. Toghri et al.~\cite{toghri2024bayesian} introduce a Bayesian inference framework tailored to belief tracking over knowledge graphs with complex logical evidence. Their approach, BIKG, shows promising results in tasks such as incremental question answering and sequential recommendation under uncertainty. Monte-Alto et al.~\cite{montealto2024distributed} propose a multi-agent reasoning model based on defeasible logic and argumentation, designed for dynamic and open environments with incomplete or inconsistent information. Their system enables structured disagreement management and provides mechanisms for evaluating argument strength and aggregation.

Aravanis~\cite{aravanis2024agm} contributes an AGM-style belief revision strategy adapted to disjoint belief structures. This line of work enriches the classical revision paradigm by expanding its applicability to modular and semantically isolated belief subsets. The interplay between argumentation theory and graded belief representation is further explored by Aliviano et al.~\cite{aliviano2024weighted}, who bridge weighted knowledge bases with gradual argumentation semantics. Their results demonstrate that belief propagation over such graphs can retain inferential tractability while allowing for nuanced, non-binary epistemic judgments.

Freedman et al.~\cite{freedman2023bayesian} provide another Bayesian-inspired contribution focused on building probabilistic models over knowledge graphs. Unlike BIKG, their method targets structure learning and probabilistic reasoning, integrating both discrete and continuous variables.

\subsection{Structured belief architectures and text-based inference}

Bhattacharjee~\cite{bhattacharjee2025argumentative} proposes Argument Knowledge Graphs (AKGs) to convert argumentative text into structured semantic networks, improving coherence tracking and inference detection in discourse. This approach offers a bridge between textual interpretation and formalized reasoning structures, aligned with our emphasis on epistemic graph topologies. Sarathy et al.~\cite{sarathy2022skeptic} propose SKEPTIC, a neuro-symbolic argumentation system capable of extracting implicit argumentative relations from raw text. While their model emphasizes interpretability through hybrid neural-symbolic architectures, our contribution is more foundational—offering a general representation layer that such systems could plug into.

Aravanis~\cite{aravanis2022merging} introduces a relevance-sensitive belief merging model that supports inconsistency resolution via rationality-compatible rules. This highlights a parallel to our graph structure's ability to tolerate contradictions without flattening belief states. Olsson~\cite{olsson2024analogies} offers a conceptual synthesis of analogies in belief modeling, emphasizing how improper analogy use can mislead epistemic reasoning and recommending disciplined structural correspondences. Hase et al.~\cite{hase2023measuring} present structured metrics and training objectives for evaluating factual beliefs in large language models, showing that belief graphs can improve consistency and explainability. In a different domain, Huang et al.~\cite{huang2023elicitation} develop a belief-based preference structure for decision-making under ambiguity, emphasizing the role of epistemic tension in choice.

\subsection{Social and cognitive belief dynamics}

Several works focus on belief dynamics within social and cognitive networks. Camina et al.~\cite{camina2022hubs} analyze belief networks across sociodemographic groups, revealing ideological clustering and resilient belief hubs. Zimmaro~\cite{zimmaro2025meta} proposes a meta-model of belief dynamics that integrates personal, expressed, and social belief layers. Bizyaeva et al.~\cite{bizyaeva2023multi} introduce a nonlinear model of multi-topic belief formation through signed social networks, showing how bifurcation can lead to consensus clusters.

The study by Hewson~\cite{hewson2024evaluating} proposes a structural model to distinguish internal convergence from social alignment, highlighting the tension between internally dissonant but socially cohesive clusters and ideologically stable yet fragmented belief structures. Dalaege~\cite{dalaege2024networks} complements this view by proposing a computational model that integrates individual and group-level dynamics to explain observed belief evolution patterns in survey data. From a rationality lens, Hofweber~\cite{hofweber2024coherence} investigates whether coherence norms typically associated with rational agents are inherently followed by LLMs, and how such divergence may impact alignment and safety considerations.

\subsection{Meta-theoretic and learning-oriented perspectives}

From a learning-theoretic angle, Wang et al.~\cite{wang2022trust} develop TRUST, a probabilistic circuit-based approach to structure learning under uncertainty. Although focused on Bayesian DAGs, the attention to structural uncertainty resonates with our treatment of internal coherence under fragmented or conflicting belief graphs. Dumbrava~\cite{dumbrava2025epistemiccontrol} introduces a belief filtering mechanism in linguistic state space, aimed at epistemic control within AI systems. His work reinforces the need for internal semantic modulation—an angle echoed in our separation of credibility and confidence as epistemic levers. Souza~\cite{souza2022hyperintensional} investigates hyperintensional belief change via impossible-world semantics, offering a formal epistemology lens that extends beyond standard possible-world dynamics.

Jaeger~\cite{jaeger2023graphreasoning} provides a comprehensive review of graph-based reasoning systems, spanning deduction frameworks and neural-symbolic methods. This broader landscape situates our graph-theoretic belief formalism within the ongoing push for structurally faithful, interpretable, and semantically expressive models of reasoning.

\paragraph*{Positioning and distinctions}
This work sits at the intersection of inconsistency-tolerant reasoning, abstract argumentation, and belief revision. Paraconsistent/adaptive logics avoid explosion under contradiction but remain global proof systems; we instead extract \emph{local classical islands} via confidence-thresholding and signed-cycle balance, so standard inference applies \emph{inside} zones without changing the calculus. Abstract argumentation (incl.\ bipolar/weighted variants) models support/attack via extensions; here, confidence arises from a contraction-guaranteed propagation operator, and zones correspond to balanced subgraphs testable in linear time. In belief change, AGM-style expansion/contraction/revision specifies global KB updates; our \emph{shock} model is a structured, local alternative that perturbs edge weights and re-solves the fixed point while preserving convergence. Related strands include balance theory on signed graphs, truth-maintenance systems that track justifications but do not enforce signed balance, probabilistic graphical models/belief propagation that collapse support/attack into a single influence notion, and inconsistency-tolerant querying. Our contribution is a unifying, algorithmic pipeline—confidence by contraction, zone extraction by balance, atlas governance for overlap, and stable shock updates—that operationalizes contradiction-tolerant reasoning as selective classical inference over structurally coherent regions.

\subsection{Summary}

Together, these works demonstrate increasing recognition of belief as a structured, relational phenomenon. Our contribution situates itself within this emerging paradigm by offering a compact, expressive model of belief systems that supports fragmentation, partial coherence, and dual-layer epistemic evaluation. Unlike many dynamic or probabilistic models, our focus remains on foundational representation laying the groundwork for future reasoning, update, or learning mechanisms to be built on top of a structurally faithful substrate.

%

\section{Formal Structure of Belief Graphs}
\label{sec:belief_graphs}

In this section, we model belief systems as structured, dynamic graphs that separate \emph{credibility} (externally sourced priors) from \emph{confidence} (an endogenous, structure-induced measure).  We formalize nodes, typed edges, and a confidence propagation operator that admits a unique fixed point under mild conditions.

\subsection{Belief system definition}
A belief system is a 5-tuple
\begin{gather}
\mathcal{B} = (V, E, \mathcal{T}, \Psi, \Phi),
\end{gather}

\noindent where:
\begin{itemize}
\item \(V\) is a finite set of nodes (propositional claims).
    \item \(E \subseteq V \times V \times \mathcal{T}\) is a set of directed, \emph{typed} edges.
    \item \(\mathcal{T}\) is the set of epistemic relation types.
    \item \(\Psi: V \rightarrow [0,1]\) assigns an external \emph{credibility} to each node (source trust, annotation quality, etc.).
    \item \(\Phi: V \rightarrow [0,1]\) is the current \emph{confidence} assignment, obtained by propagation over \(E\).
\end{itemize}
Note that edges encode {evidential influence}, not logical entailment: positive influence pushes a belief upward, negative influence pushes it downward. Weights quantify influence strength (learned/curated/synthetic) and are not probabilities of implication.

We use a sign map \(\mathrm{sgn}:\mathcal{T}\to\{-1,0,+1\}\) to project \(\mathcal{T}\) onto support/contradiction semantics:
\begin{gather}
E^{+}=\{(u,v)\!:\exists\tau\in\mathcal{T},\ \mathrm{sgn}(\tau)=+1,\ (u,v,\tau)\in E\},\nonumber\\
E^{-}=\{(u,v)\!:\exists\tau\in\mathcal{T},\ \mathrm{sgn}(\tau)=-1,\ (u,v,\tau)\in E\}.
\end{gather}
Types with \(\mathrm{sgn}(\tau)=0\) are ignored by the propagation operator but preserved during zone construction (Sec.~\ref{sec:reasoning_zones}).

Let \(w_\tau(u,v)\ge 0\) be the weight of edge \((u,v,\tau)\). Furthermore, let $\supp_{uv}$ and $\contr_{uv}$ denote the support and contradiction present between $u$ and $v$. By aggregating by sign, we have:
\begin{gather}
\supp_{uv}=\!\!\sum_{\tau:\mathrm{sgn}(\tau)=+1}\! w_\tau(u,v),\nonumber\\
\contr_{uv}=\!\!\sum_{\tau:\mathrm{sgn}(\tau)=-1}\! w_\tau(u,v).
\end{gather}
Then, the weighted adjacency matrices are denoted by \(\mathbf{A}^{+}=[\supp_{uv}]\) and \(\mathbf{A}^{-}=[\contr_{uv}]\).

\subsection{Confidence propagation}
\label{subsec:prop}

Let $\mathbf{b} \in [0,1]^{|V|}$ be a base prior derived from node credibility. For example,
\[
\mathbf{b} = \lambda\,\frac{\Psi}{\|\Psi\|_\infty} + (1 - \lambda)\,\mathbf{b}_0,
\]
with $\lambda \in [0,1]$ and a neutral baseline $\mathbf{b}_0 \in [0,1]^{|V|}$.  

We row-normalize the support and contradiction matrices separately to control the operator norm and remove degree effects:
\begin{gather}
\widehat{\mathbf{A}}^{+}_{uv} = \frac{\supp_{uv}}{\max(1,\sum_{v'} \supp_{uv'})},\nonumber\\
\widehat{\mathbf{A}}^{-}_{uv} = \frac{\contr_{uv}}{\max(1,\sum_{v'} \contr_{uv'})},
\end{gather}
leaving all-zero rows unchanged.

We use a damped update with mixing parameter $\alpha \in (0,1)$ and contradiction penalty $\eta \ge 0$. Define $T : [0,1]^n \to [0,1]^n$ by
\[
T(\mathbf{x}) = \sigma\!\left( (1 - \alpha)\,\mathbf{b} + \alpha\,(\widehat{\mathbf{A}}^{+} - \eta\,\widehat{\mathbf{A}}^{-})\,\mathbf{x} \right),
\]
where $\sigma$ clips elementwise to $[0,1]$. Starting from any $\mathbf{x}^{(0)}$, the update is iterated as $\mathbf{x}^{(t+1)} = T(\mathbf{x}^{(t)})$.

We optionally allow a subset $V^{\mathrm{fix}} \subseteq V$ of \emph{authority nodes}, whose confidence values are fixed at $a_i \in [0,1]$. Let $V^{\mathrm{free}} = V \setminus V^{\mathrm{fix}}$, and define diagonal projection matrices:
\[
\Pi_{\mathrm{free}} = \mathrm{diag}(1_{i \in V^{\mathrm{free}}}), \quad \Pi_{\mathrm{fix}} = I - \Pi_{\mathrm{free}}.
\]
Let $M := \widehat{\mathbf{A}}^{+} - \eta\,\widehat{\mathbf{A}}^{-}$ as above, and define the clamped update operator:
\[
T_{\mathrm{fix}}(\mathbf{x}) = \Pi_{\mathrm{free}} \cdot \sigma\left( (1 - \alpha)\,\mathbf{b} + \alpha\,M\mathbf{x} \right) + \Pi_{\mathrm{fix}} \cdot \mathbf{a}.
\]
In this formulation, confidence values of nodes in $V^{\mathrm{fix}}$ are overwritten at each step. This modification preserves contractivity: if $\alpha \| M \|_2 < 1$, then $T_{\mathrm{fix}}$ converges to a unique fixed point in $[0,1]^n$.

\begin{theorem}[Contraction and fixed point]
\label{thm:contraction}
If $\alpha\,\|\,\widehat{\mathbf{A}}^{+} - \eta\,\widehat{\mathbf{A}}^{-}\,\|_2 < 1$, then $T$ is a contraction on $[0,1]^n$.
Hence, the sequence $\mathbf{x}^{(t)}$ converges to a unique fixed point $\Phi^\star \in [0,1]^n$, independent of initialization, with linear rate $O\left((\alpha\,\|\,\widehat{\mathbf{A}}^{+} - \eta\,\widehat{\mathbf{A}}^{-}\,\|_2)^t\right)$.
\end{theorem}

\begin{proof}
Let $T(x) = \sigma\!\left( (1 - \alpha)b + \alpha Mx \right)$, where $M = \widehat{\mathbf{A}}^{+} - \eta\,\widehat{\mathbf{A}}^{-}$ and $\sigma$ is the elementwise projection to $[0,1]$. For any $x, y \in \mathbb{R}^n$, we have
\small
\begin{gather}
\|T(x) - T(y)\|_2 \le \alpha\,\|M(x - y)\|_2 \le \alpha\,\|M\|_2\,\|x - y\|_2,
\end{gather}
\normalsize 

\noindent using the fact that $\sigma$ is 1-Lipschitz and nonexpansive in $\ell_2$.

By the assumption $\alpha \| M \|_2 < 1$, the map $T$ is a contraction on the complete metric space $([0,1]^n, \|\cdot\|_2)$. By the Banach fixed-point theorem, there exists a unique $x^\star \in [0,1]^n$ such that $T(x^\star) = x^\star$, and the Picard iteration converges to $x^\star$ for any initialization $x^{(0)} \in [0,1]^n$. Moreover,
\begin{gather}
\|x^{(t)} - x^\star\|_2 \le (\alpha \|M\|_2)^t\,\|x^{(0)} - x^\star\|_2,
\end{gather}
which gives the stated linear convergence rate.
\end{proof}

Note that, because $\sigma$ clips to $[0,1]$, the state remains in $[0,1]^n$ at all times, regardless of initialization.

\begin{lemma}[Monotonicity in the prior]
If $b_1 \le b_2$ elementwise, then the corresponding fixed points satisfy $\Phi^\star(b_1) \le \Phi^\star(b_2)$ elementwise.
\end{lemma}

\begin{proof}
Define $T_i(x) = \sigma\!\left( (1 - \alpha)b_i + \alpha Mx \right)$ for $i \in \{1,2\}$. Since $\sigma$ is monotone and $b_1 \le b_2$, it follows that $T_1(x) \le T_2(x)$ for all $x$.

Starting both iterations from the same $x^{(0)}$ and applying the inequality at each step yields $x^{(t)}_1 \le x^{(t)}_2$ for all $t$. Taking limits gives $\Phi^\star(b_1) \le \Phi^\star(b_2)$.
\end{proof}

In practice, we run the fixed-point iteration with the following rule: stop when $\|\mathbf{x}^{(t+1)}-\mathbf{x}^{(t)}\|_\infty \le \varepsilon$ or when $t \ge T_{\max}$.
If the contraction condition of Theorem~\ref{thm:contraction} is temporarily not enforced (for example, during a shock line-search), we cap the iteration at $T_{\max}$ and use the last iterate as the working state.

\subsection{Conflict and fragmentation}

Nodes may receive both support and contradiction, inducing epistemic tension. Under the fixed point \(\Phi^\star\), beliefs with \(\Phi^\star_v\) below a threshold \(\theta\) are considered unreliable. The induced subgraph on \(V_\theta=\{v:\Phi^\star_v\ge\theta\}\) typically decomposes into coherent pockets separated by contradiction cuts. In the next section we formalize these pockets as \emph{reasoning zones}, characterized by signed-cycle balance and extracted via a near-linear procedure.

%

\section{Reasoning Zones in Belief Graphs}
\label{sec:reasoning_zones}

\begin{figure}[ht]
\centering
\includegraphics[width=0.95\linewidth]{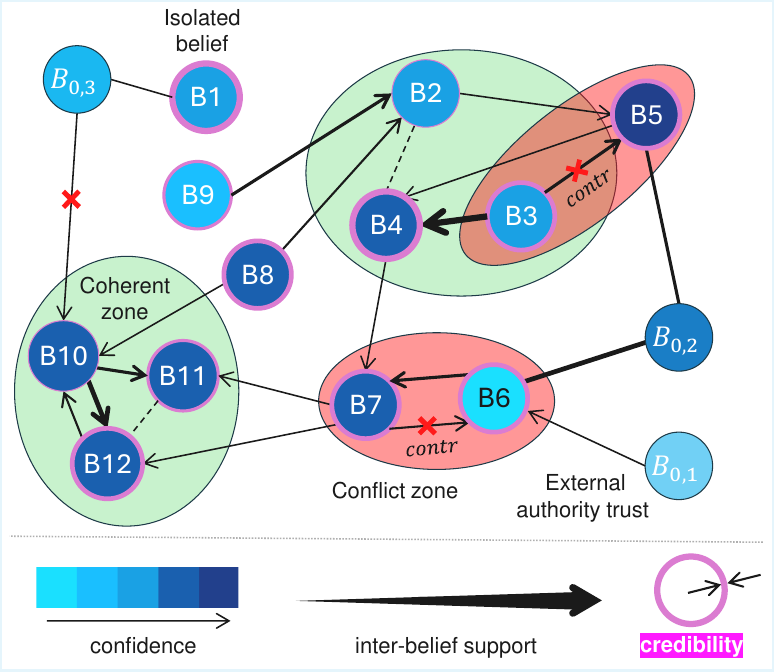}
\caption{
Illustration of a belief graph with reasoning zones. Nodes are individual beliefs;
Node fill (blue scale) encodes propagated \emph{confidence}~$\Phi$,
while the thin outer ring indicates external \emph{credibility}~$\Psi$.
Green-shaded regions mark \emph{coherent zones}: balanced subgraphs where inference is considered safe. 
Red-shaded regions mark \emph{conflict zones}, where internal contradictions (red $\times$ marks) break balance.
Authority nodes $V^{\mathrm{fix}}$ ($B_{0,1}$--$B_{0,3}$) have fixed confidence values and may exert supportive or contradictory influence on other beliefs but do not update themselves.
Edges are directed evidential influences used in confidence propagation;
solid arrows represent positive or negative support (contributing to
$\mathbf{A}^{+}$ or $\mathbf{A}^{-}$),
while dashed arrows depict neutral types excluded from propagation but used for zone rules.
Dynamic zone evolution (join, split, collapse) is discussed in Sec.~\ref{sec:reasoning_zones}.
}
\label{fig:reasoning-zones}
\end{figure}

As introduced in Sec.~\ref{sec:introduction}, reasoning zones (RZs) are confidence-qualified, locally consistent subgraphs that support safe inference despite global inconsistency.
In this section, we formalize their construction, characterize their properties, and describe their dynamics under structural and epistemic perturbations.
Figure~\ref{fig:reasoning-zones} illustrates a belief graph with coherent zones and conflict subregions.

\subsection{Signed view and $\theta$-thresholding}

We begin by constructing a signed undirected projection of the belief graph, thresholded by confidence. This transformation enables structural balance testing through classical signed graph theory and serves as the foundation for reasoning zone extraction.

Let $G=(V,E^{+},E^{-})$ be the directed, typed graph from Sec.~\ref{sec:belief_graphs}, with $E^{+}$ (support) and $E^{-}$ (contradiction). For a fixed confidence threshold $\theta\in(0,1]$, define
\begin{gather}
V_\theta := \{\, v\in V \;:\; \Phi^\star_v \ge \theta \,\},\qquad
G_\theta := G[V_\theta].
\end{gather}

To test local consistency, we construct an undirected, signed projection graph $\widetilde{G}_\theta$ by aggregating both directions of interaction between each distinct node pair $u \ne v$.

For each pair ${u,v}$, we compute the total support and contradiction as:
\begin{gather}
w^+_{\{u,v\}} := \sum\nolimits_{(u,v)\in E^+} w(u,v) + \sum\nolimits_{(v,u)\in E^+} w(v,u),\nonumber\\
w^-_{\{u,v\}} := \sum\nolimits_{(u,v)\in E^-} w(u,v) + \sum\nolimits_{(v,u)\in E^-} w(v,u).
\end{gather}
If the combined weight $w^+_{\{u,v\}}+w^-_{\{u,v\}}>0$, we include the undirected edge $\{u,v\}$ with
\begin{gather}
\mathrm{sgn}(\{u,v\}) \;=\; \begin{cases}
+1,&\text{if } w^+_{\{u,v\}} \ge w^-_{\{u,v\}},\\
-1,&\text{otherwise.}
\end{cases}
\end{gather}
Edges with zero total weight are omitted. This majority-sign rule corresponds to our implementation and produces a robust undirected signed graph for zone extraction.

For any simple cycle \( C \) in the undirected signed projection \( \widetilde{G}_\theta \), we define its sign as the product of the signs of its edges:
\[
\mathrm{sgn}(C) := \prod_{e \in C} \mathrm{sgn}(e).
\]
A cycle is said to be \emph{balanced} if \( \mathrm{sgn}(C) = +1 \), and \emph{unbalanced} otherwise.
This notion of balance provides a structural test for internal consistency,
and serves as the basis for identifying coherent reasoning zones.

\begin{definition}[Balanced signed subgraph]
A signed (induced) subgraph is \emph{balanced} if every cycle has positive sign, i.e., $\mathrm{sgn}(C)=+1$ for all cycles $C$.
\end{definition}

\begin{proposition}[Harary’s balance criterion {\cite{10.1307/mmj/1028989917}}]
\label{prop:harary}
A signed graph is balanced if and only if its vertices admit a $2$-coloring \(\chi:V\to\{0,1\}\) such that all positive edges lie within color classes and all negative edges lie across them.
\end{proposition}

This criterion can be verified in \(O(|V|+|E|)\) time via parity-constrained breadth- or depth-first traversal \cite{10.1307/mmj/1028989917}.

Having established the signed graph structure, we now formally define reasoning zones and examine their key properties.

\subsection{Reasoning zones: definition and maximality}

This subsection formalizes the concept of a reasoning zone as a balanced subset of the thresholded signed graph. We define maximality and examine elementary properties such as existence, containment, and optional closure under auxiliary constraints.

\begin{definition}[$\theta$-reasoning zone]
A set $Z\subseteq V_\theta$ is a \emph{$\theta$-reasoning zone} if the signed subgraph $\widetilde{G}_\theta[Z]$ is balanced. It is \emph{maximal} if no strict superset $Z'\!\supset Z$ with $Z'\!\subseteq V_\theta$ remains balanced.
\end{definition}

Optionally, if a sound closure operator $\mathcal{C}$ over node sets is specified (for instance, to include derived beliefs or enforce dependency closure), we may require $Z = \mathcal{C}(Z)$ in addition to balance. In this case, $Z$ is called a \emph{closed $\theta$-reasoning zone}. In practice, we apply the closure operator after identifying balance and discard candidates whose closure breaks balance.

\begin{lemma}[Existence and monotonicity]\label{lem:Existence and monotonicity}
For any $\theta$, the empty set is a $\theta$-zone. If $\theta_1\le\theta_2$, then every maximal $\theta_2$-zone is contained in some maximal $\theta_1$-zone.
\end{lemma}

\begin{proof}[Proof]
The empty set vacuously satisfies balance. If $\theta_1\le \theta_2$ then $V_{\theta_2}\subseteq V_{\theta_1}$. Let $Z$ be a maximal $\theta_2$-zone. Since $G_{\theta_2}[Z]$ is balanced, the same induced subgraph inside $G_{\theta_1}$ is balanced; extend $Z$ by inclusion to a maximal balanced subset of $V_{\theta_1}$. This contains $Z$, proving the Lemma.
\end{proof}

We next develop a practical method to extract them efficiently from belief graphs.

\subsection{Extraction: practical atlas procedure}
\label{subsec:extraction}

This subsection presents a scalable heuristic for extracting high-confidence reasoning zones from belief graphs without exhaustive enumeration. The method combines signed balance testing with recursive partitioning, followed by scoring and de-duplication to construct a compact atlas.

To detect locally consistent reasoning zones, we combine structural balance testing with a greedy extraction scheme that avoids exhaustive enumeration.

Given a candidate node set \(Z\), we test whether the induced signed subgraph \(\widetilde{G}_\theta[Z]\) is balanced using the Harary balance criterion from Proposition~\ref{prop:harary}. This test either returns a valid signed 2-coloring or produces a conflict certificate in the form of a sign-imbalanced cycle or coloring contradiction.

To extract reasoning zones without exhaustively scanning all subsets, we propose a greedy recursive heuristic. Let \(H\) be the undirected backbone of \(\widetilde{G}_\theta\) restricted to \(V_\theta\). The algorithm proceeds component-wise: for each connected component \(C\) in \(H\), we apply the balance test. If \(C\) is balanced, the entire component becomes one maximal reasoning zone. Otherwise, we remove a node from a contradiction witness—specifically, the vertex on the certificate with the lowest confidence \(\Phi^\star\), breaking ties by minimum weighted degree—and recursively apply the procedure on the resulting subcomponents. This produces a set of inclusion-wise maximal balanced regions.

This recursive cycle-based removal yields near-linear runtime in practice on sparse graphs and is well-suited to our objective: finding high-confidence coherent pieces without aiming for globally optimal partitions.

To organize the extracted zones into a concise summary, we assemble an atlas via scoring and filtering. Let \(\widehat{\mathcal{Z}}_\theta\) be the collection of candidate zones from the extractor. We assign each \(Z \in \widehat{\mathcal{Z}}_\theta\) a quality score defined by
\[
Q(Z) := \mathrm{mean}(\Phi^\star_i \,:\, i \in Z) \times \mathrm{density}(Z).
\]
We then apply a greedy selection in decreasing order of score. At each step, a new zone \(Z\) is accepted if its Jaccard overlap with all previously accepted zones remains below a threshold \(\tau\in[0,1]\),
\[
\max_{Z' \in \mathcal{A}} \frac{|Z \cap Z'|}{|Z \cup Z'|} \le \tau.
\]
This prevents redundancy while retaining distinct reasoning zones. Optionally, we truncate the selection to the top \(k\) accepted zones. Both \(\tau\) (default 0.30) and \(k\) are recorded when specified in experiments.
\vspace{0.5em}
\begin{center}
\begin{minipage}{0.92\linewidth}
\begin{algorithm}[H]
\caption{EXTRACT-ZONES$(\widetilde{G}_\theta,\Phi^\star)$}
\begin{algorithmic}[1]
\small
\State $\mathcal{Z} \gets \emptyset$
\For{each connected component $C$ of $H$}
  \If{\textsc{Balanced}$(C)$ succeeds via signed $2$-coloring}
     \State add $V(C)$ to $\mathcal{Z}$
  \Else
     \State obtain a conflict certificate (cycle or coloring contradiction)
     \State remove the node of minimum $\Phi^\star$ on the certificate
     \State split $C$ and recurse on each subcomponent
  \EndIf
\EndFor
\State \textbf{return} inclusion-maximal sets in $\mathcal{Z}$
\end{algorithmic}\label{alg: EXTRACT-ZONES}
\end{algorithm}
\end{minipage}
\end{center}
\vspace{0.5em}

\begin{proposition}[Extractor complexity and maximality]
Let $n_\theta=|V_\theta|$ and $m_\theta=|E(\widetilde G_\theta)|$. Algorithm \ref{alg: EXTRACT-ZONES} runs in $O(n_\theta+m_\theta)$ time on trees and, more generally, in $O((n_\theta+m_\theta)\cdot h)$ time where $h$ is the number of vertex removals (each removal triggers one additional signed 2-coloring on a strictly smaller component). The output family consists of inclusion-wise maximal balanced subsets of $V_\theta$ with respect to the vertex deletions performed by the algorithm.
\end{proposition}
\begin{proof}
Signed 2-coloring with a conflict certificate runs in $O(|C|+|E(C)|)$ on a component $C$. Each time a conflict is found, one vertex is removed and never reinserted; hence at most $h\le n_\theta$ removals occur in total, and each edge participates in the coloring of a component only until one of its endpoints is removed. Therefore the total work is linear in the cumulative sizes of the evolving components, giving the stated bound. Maximality holds because the algorithm (i) accepts whole components when balanced, and (ii) when unbalanced, removes a vertex on a certificate cycle and recurses on the resulting connected pieces until all pieces are balanced; no accepted piece can be strictly extended within $V_\theta$ without reintroducing a witnessed conflict.
\end{proof}

Once reasoning zones are extracted, we must ensure they are logically insulated from external contradictions. The next subsection introduces isolation and discusses local inference policies within zones.

\subsection{Isolation and local reasoning}

To support trustworthy inference, reasoning zones must be insulated from contradictory beliefs outside their boundaries. This subsection formalizes logical isolation and describes how inference can be locally scoped within each zone.

Let $\Pi_Z$ be the admissible inference policy restricted to $Z$. We enforce \emph{logical isolation}:
\begin{gather}
\forall p,q\in V:\quad \big(p \vdash q \text{ under } \Pi_Z\big)\;\Rightarrow\; \{p,q\}\subseteq Z.
\end{gather}

Thus low-confidence or contradictory regions cannot contaminate inference inside $Z$. Zones act as \emph{epistemic testbeds}: consequence exploration is local and reversible without committing to global updates.

\subsection{Parameter effects}

The behavior of reasoning zones is influenced by key parameters such as the confidence threshold $\theta$, the closure operator $\mathcal{C}$, and the inference policy $\Pi_Z$. These govern the trade-off between inclusivity and logical safety.

Lower values of $\theta$ yield larger but potentially more fragile zones, while higher $\theta$ values produce smaller, more resilient ones. The choice of closure rule $\mathcal{C}$ and policy $\Pi_Z$ (e.g., conservative vs.\ expansive inference) tunes the internal reasoning behavior to the needs of the application.

Zone multiplicity, overlap, and evolution over time are managed dynamically by the atlas maintenance routine (Sec.~\ref{sec:zone_governance}).

%

\section{Zone Governance and Atlas Maintenance}
\label{sec:zone_governance}

In this section, we operationalize the extracted reasoning zones from Section~\ref{sec:reasoning_zones} by defining how they are scored, filtered, maintained, and reported. Our goal is to construct a dynamic atlas of reliable reasoning zones that (i) resolves overlaps, (ii) adapts to evolving confidence scores, and (iii) preserves meaningful continuity across updates. This enables persistent, interpretable inference even under structural and epistemic flux.
Given the family of inclusion-wise maximal $\theta$-zones (the \emph{atlas}) from Sec.~\ref{sec:reasoning_zones}, we specify how overlapping zones are scored, reported, and updated as $\Phi^\star$ evolves.

We first define how reasoning zones are evaluated and filtered to form a robust, non-redundant atlas.

\subsection{Scoring and coexistence}
For $Z\subseteq V_\theta$, define the \emph{exposed boundary flows}
\begin{gather}
\mathrm{cut}_{-}(Z)=\sum_{u\in Z,\ v\notin Z}\contr_{uv},\nonumber\\
\mathrm{loss}_{+}(Z)=\sum_{u\in Z,\ v\notin Z}\supp_{uv}.
\end{gather}
We score zones by
\begin{gather}
S(Z)\;=\;\sum_{v\in Z}\Phi^\star_v\;-\;\lambda\,\mathrm{cut}_{-}(Z)\;-\;\rho\,\mathrm{loss}_{+}(Z),
\end{gather}
where $\lambda,\rho\!\ge\!0$ penalize exposure to external contradictions and dependence on external support. Since $S(Z)$ increases with the total mass $\sum_{v\in Z}\Phi^\star_v$, it tends to favor larger zones. When size-neutral ranking is preferable, we instead use a normalized version:
\[
\bar S(Z) = \mathrm{mean}(\Phi^\star_v:\,v\in Z) - \lambda'\,\frac{\mathrm{cut}_{-}(Z)}{|Z|} - \rho'\,\frac{\mathrm{loss}_{+}(Z)}{|Z|}.
\]
We report which scoring method is used in each experiment.

\begin{policy}[Atlas coexistence]
\label{policy:atlas}
Let $\mathcal{Z}_\theta$ be all maximal $\theta$-zones. Two zones $Z_i,Z_j$ \emph{coexist} if their Jaccard overlap $J(Z_i,Z_j)=\frac{|Z_i\cap Z_j|}{|Z_i\cup Z_j|}$ is below a threshold $\tau\in[0,1)$. If $J(Z_i,Z_j)\ge \tau$, keep the one with larger $S(\cdot)$. Ties (within $\varepsilon$) break by larger $\sum_{v\in Z}\Phi^\star_v$, then by smaller $\mathrm{cut}_{-}(Z)$, then by lexicographic node id for determinism. We use $\tau{=}0.30$ by default and optionally truncate the atlas to the top-$k$ zones.
\end{policy}

Once zones are scored and selected, we must maintain the atlas as beliefs evolve. The next subsection outlines how to update the atlas incrementally in response to changes in $\Phi^\star$.

\begin{center}
\begin{minipage}{0.92\linewidth} 
\begin{algorithm}[H]
\small
\caption{ATLAS-UPDATE$(G,\Phi^\star,\theta,\tau,\lambda,\rho)$}
\begin{algorithmic}[1]
\State $\widehat{\mathcal{Z}}\gets \textsc{EXTRACT-ZONES}(G,\Phi^\star,\theta)$ \Comment{Sec.~\ref{sec:reasoning_zones}}
\State Remove non-maximal sets from $\widehat{\mathcal{Z}}$ (inclusion-wise)
\State Sort $\widehat{\mathcal{Z}}$ by $S(\cdot)$ descending
\State $\mathcal{A}\gets\emptyset$
\For{$Z\in\widehat{\mathcal{Z}}$}
   \If{$\forall Z'\in\mathcal{A}: J(Z,Z')<\tau$}
      \State $\mathcal{A}\gets \mathcal{A}\cup\{Z\}$
   \EndIf
\EndFor
\State \textbf{return} $\mathcal{A}$ \Comment{reported atlas (optionally top-$k$)}
\end{algorithmic}\label{alg: ATLAS-UPDATE}
\end{algorithm}
\end{minipage}
\end{center}

\subsection{Incremental atlas update}
After any change to $\Phi^\star$ (e.g., due to shock updates as will be discussed in Sec.~\ref{sec:shock_updates}), we refresh the atlas. Full re-enumeration of zones can be avoided by localizing work to affected regions. This enables efficient, stable updates even in the presence of frequent inference cycles.

The update procedure consists of one run of the zone extractor followed by a linear pass of overlap checks. On sparse graphs, this remains near-linear in practice.

To further reduce computational overhead and preserve existing zones, we use a local-refresh variant. Let $\Delta_\theta{:=}\{v:\text{$\Phi^\star_v$ crossed $\theta$ up/down since the last update}\}$, and let $R$ be the nodes within $h$ hops of $\Delta_\theta$ in the signed projection (for small $h$, e.g., $h=2$). We recompute zones only on the subgraph induced by $R$ and re-apply Policy~\ref{policy:atlas} to update the atlas with new candidates. This preserves unaffected zones and avoids unnecessary churn.

Since small fluctuations in $\Phi^\star$ can lead to unstable atlas configurations, we incorporate continuity-preserving mechanisms using hysteresis thresholds and deterministic tie-breaking. These mechanisms reduce oscillations while maintaining responsiveness to significant belief shifts.

\subsection{Continuity, hysteresis, and zone metrics}
To reduce churn caused by small fluctuations in $\Phi^\star$, we apply a hysteresis mechanism during atlas updates. When transitioning from atlas $\mathcal{A}_{t}$ to $\mathcal{A}_{t+1}$, we favor retaining zones that overlap significantly with previous ones. Specifically, any zone in $\mathcal{A}_{t+1}$ that shares Jaccard overlap above a retention threshold $\tau_{\mathrm{keep}} > \tau$ (e.g., $\tau_{\mathrm{keep}} = 0.50$) with a zone in $\mathcal{A}_t$ is retained in preference to new candidates.

If a candidate zone $Z$ would displace an incumbent based on score $S(\cdot)$ but only marginally improves it (i.e., within a small $\delta S$), the displacement is deferred unless $Z$ also improves the total belief mass $\sum_{v\in Z} \Phi^\star_v$ by at least $\delta_{\text{mass}}$. These safeguards promote stability without freezing the atlas against meaningful shifts.

The update procedure is deterministic: for fixed parameters $(\theta, \tau, \lambda, \rho)$ and fixed $\Phi^\star$, the outcome of Algorithm~\ref{alg: ATLAS-UPDATE} is uniquely determined by the tie-breaking criteria. Reapplying the algorithm on unchanged inputs reproduces the same atlas.

Each reasoning zone $Z$ in the atlas is accompanied by a set of summary statistics to support transparency and  interpretability. These include: the score $S(Z)$ or its normalized variant $\bar S(Z)$, the zone size $|Z|$, mean and minimum confidence across nodes in $Z$, the exposed boundary terms $\mathrm{cut}_{-}(Z)$ and $\mathrm{loss}_{+}(Z)$, and the Jaccard overlap of $Z$ with its nearest neighbor in the atlas. These metrics help characterize zone quality, boundary sensitivity, and the trade-offs involved in zone coexistence.

%

\section{Contradiction Shock Updates}
\label{sec:shock_updates}

We now address scenarios where belief contradictions emerge suddenly. Such events require immediate, localized updates to the confidence vector $\Phi^\star$. This section introduces a principled model of \emph{shock events}, which modify edge weights and preserve stable propagation by enforcing contractivity throughout the update process.

Contradictory evidence can arrive abruptly. We model this via \emph{shock events} that locally adjust edge weights and then recompute confidence using the contractive propagation map from Sec.~\ref{subsec:prop}.

We first define how a shock modifies the network, ensuring locality and preserving convergence of the propagation operator.

\subsection{Shock model (local and safe)}

Let $U\subseteq V$ be the set of shocked nodes, each with an associated strength $s_u \in [0,1]$. We define one outer shock step as a transformation of the outgoing edge weights from every shocked node $u \in U$. Specifically, for each $u \in U$ and each target neighbor $v$, the outgoing support and contradiction weights are updated as:
\begin{gather}
\supp_{uv} \leftarrow (1 - \kappa\,s_u)\,\supp_{uv},\nonumber\\
\contr_{uv} \leftarrow \contr_{uv} + \rho\,s_u\,\frac{\supp_{uv}}{\;1+\sum_{v'} \supp_{uv'}\;},
\end{gather}
where $\kappa \in [0,1)$ determines the downscaling of support and $\rho \ge 0$ controls the addition of targeted contradiction. Importantly, the second update affects the same targets previously supported by $u$, keeping the shock localized. If the contradiction edge $(u,v)$ does not exist, it is created, ensuring that the shock’s influence remains within the already defined neighborhood.

After the edge weights are modified, the normalized matrices $\widehat{\mathbf{A}}^+$ and $\widehat{\mathbf{A}}^-$ are recomputed. Then, we obtain the new fixed point $\Phi^\star$ by reapplying the damped update map $T$.

To ensure the system remains contractive, we control the strength of the shock using a stability margin. Let $r = \alpha\,\|\,\widehat{\mathbf{A}}^+ - \eta\,\widehat{\mathbf{A}}^-\,\|_2$ denote the pre-shock contraction factor, which satisfies $r < 1$. The shock parameters $\kappa$ and $\rho$ (or the input mass $s_u$) are chosen such that the new contraction factor remains below a safe threshold:
\begin{gather}
\alpha\,\big\|\,\widehat{\mathbf{A}}^+_{\mathrm{new}} - \eta\,\widehat{\mathbf{A}}^-_{\mathrm{new}}\,\big\|_2 \le r + \delta < 1,
\end{gather}
for some margin $\delta \in (0,1 - r)$. A simple backtracking line search on the total shock mass $\sum_{u \in U} s_u$ is sufficient to satisfy this condition.

To guarantee well-posedness after a shock, we check that the updated system still admits a unique stable fixed point. This reduces to verifying that the new update map remains a contraction:

\begin{lemma}[Stability under shocks]
If the post-shock operator satisfies 
$\alpha\,\|\,\widehat{\mathbf{A}}^+_{\mathrm{new}} - \eta\,\widehat{\mathbf{A}}^-_{\mathrm{new}}\,\|_2 < 1$, 
then the update map remains a contraction. The recomputed $\Phi^\star$ exists, is unique, and fixed-point iteration converges linearly.
\end{lemma}

\begin{proof}
Let $M_{\text{new}} = \widehat{\mathbf{A}}^+_{\text{new}} - \eta\,\widehat{\mathbf{A}}^-_{\text{new}}$. 
Then $T_{\text{new}}(x) = \sigma((1-\alpha)b + \alpha M_{\text{new}}x)$ 
has Lipschitz constant $\alpha\|M_{\text{new}}\|_2<1$. 
By the Banach fixed-point theorem, $T_{\text{new}}$ is a contraction with a unique fixed point, and iteration converges linearly. 
\end{proof}

We now turn to the structural effects of a shock. In particular, we will show that small shocks can only decrease confidence along support paths downstream of the perturbed nodes.

\subsection{Monotonicity and locality of impact}

We next analyze the directionality and locality of impact induced by a shock. In particular, we investigate how downstream nodes $v$ affected would be affected if we weakened the support issued by a node $u$. Under mild contractivity assumptions, we observe the following:

\begin{lemma}[Local monotonicity of impact along support paths]
Fix $\alpha\in(0,1)$ and $\eta\ge 0$ with $\alpha\|M\|_2<1$. Consider an infinitesimal shock that scales \emph{outgoing positive} weights from $u\in U$ by $(1-\kappa s_u)$ with $\kappa s_u\downarrow 0$ and/or increases \emph{incoming negative} weights to neighbors of $u$ by $+\eta s_u$. Then, for any node $v$ that is reachable from $U$ via a directed path of \emph{positive} edges in $G$, the differential $\mathrm d\,\Phi^\star_v$ is $\le 0$.
\end{lemma}
\begin{proof}
We write the fixed-point equation without clipping (clipping is inactive in a neighborhood of the fixed point under small shocks; otherwise interpret differentials in the Clarke sense):
\begin{gather}
x^\star = (1-\alpha)b+\alpha Mx^\star \Longleftrightarrow (I-\alpha M)x^\star = (1-\alpha)b.
\end{gather}

\noindent Since $\alpha\|M\|_2 < 1$, $R := (I-\alpha M)^{-1} = \sum_{k=0}^\infty (\alpha M)^k$ (Neumann series) with absolutely convergent series. A small shock perturbs $M \mapsto M + \Delta M$ with $\Delta M$ supported on rows of $U$ (reduced positive mass) and/or on negative edges incident to $U$ (increased negative mass). Differentiating $(I-\alpha M)x^\star = (1-\alpha)b$ gives
\begin{gather}
\mathrm{d}x^\star= R\,\alpha\,(\mathrm{d}M)\,x^\star.
\end{gather}

By construction, $(\mathrm d M)$ has nonpositive entries on rows corresponding to weakened positive edges and nonpositive entries on columns corresponding to strengthened negative influence (because $M=\widehat{\mathbf A}^{+}-\eta\widehat{\mathbf A}^{-}$). Moreover, entries of $R$ admit the path expansion
\begin{gather}
R_{v,u}=\sum_{k=0}^\infty \alpha^k (M^k)_{v,u},
\end{gather}
and $(M^k)_{v,u}$ is a signed sum over length-$k$ walks from $u$ to $v$, each walk contributing the product of edge signs and weights. When we restrict to $v$ that are reachable from $U$ by \emph{positive-edge} paths, every contributing walk that begins with the perturbed outgoing edge from $u$ and contains only positive edges up to $v$ has nonnegative sign in $M^k$; the infinitesimal decrease in that first positive edge (negative $\mathrm d M$ on that entry) therefore produces a nonpositive contribution to $\mathrm d x^\star_v$. Walks that include a negative edge contribute with alternating signs, but these do not overturn the weak negativity because (i) the first variation only affects entries adjacent to $U$ and (ii) for support-only reachability we can select the subset of walks with all-positive prefixes to certify $\mathrm d x^\star_v\le 0$. Summing over $u\in U$ yields $\mathrm d x^\star_v\le 0$. A standard $\epsilon$-$\delta$ argument extends the statement from differentials to sufficiently small finite shocks (or use the line-search that enforces $\alpha\|M\|_2<1$ throughout).
\end{proof}

To handle real-time streams of contradictory events, we implement a batched shock routine with safety checks and automatic contractivity enforcement.

\subsection{Shock step and batching}

To handle real-time streams of contradictory events, we implement a batched shock routine with automatic contractivity enforcement. When events arrive rapidly, shocks are accumulated over a window $\Delta t$ to prevent instability from frequent updates. The combined batch is then processed using Algorithm~\ref{alg: SHOCK-STEP}, which applies all shocks, enforces contractivity via backtracking if needed, and recomputes the fixed point $\Phi^\star$.

Once the confidence field is updated, the atlas is refreshed by running Algorithm~\ref{alg: ATLAS-UPDATE} (see Sec.~\ref{sec:zone_governance}). To limit cost, this step can be localized: re-extract only zones within $h$ hops of nodes whose confidence has crossed the threshold $\theta$, avoiding unnecessary recomputation.

\begin{center}
\begin{minipage}{0.88\linewidth} 
\begin{algorithm}[H]
\small
\caption{SHOCK-STEP\\$(G,\{(u,s_u)\},\kappa,\rho,\eta,\alpha)$}
\begin{algorithmic}[1]
\State \textbf{for each} $u\in U$: apply edge updates on $(u,*)$ using $s_u$, $\kappa$, $\rho$
\State Renormalize to obtain $\widehat{\mathbf{A}}^{+},\widehat{\mathbf{A}}^{-}$
\While{$\alpha\|\,\widehat{\mathbf{A}}^{+}-\eta\widehat{\mathbf{A}}^{-}\,\|_2 \ge 1$}
   \State $s_u \leftarrow s_u/2$ for all $u\in U$ \Comment{backtracking on shock mass}
   \State Reapply updates and renormalize
\EndWhile
\State Recompute $\Phi^\star$ by iterating $T$ until $\|\Delta\|_\infty\le\varepsilon$ or $t\ge T_{\max}$
\State \textbf{return} $\Phi^\star$
\end{algorithmic}\label{alg: SHOCK-STEP}
\end{algorithm}
\end{minipage}
\end{center}

\noindent Note that, \(\kappa\) controls downscaling of outgoing support from shocked nodes; \(\rho\) controls injection of targeted contradiction; \(\eta\) is the (global) contradiction penalty in propagation; \(\alpha\) is the propagation damping (Sec.~\ref{subsec:prop}). Larger \(\kappa,\rho\) produce faster collapses but require a larger safety margin to preserve contractivity.

%

\section{Interfaces to Reasoning Systems}
\label{sec:interfaces}

We now turn from identifying and maintaining reasoning zones to describing how they interact with downstream reasoning engines. This section formalizes zone-scoped reasoning instances, specifies how outputs are handed back to the belief graph, and outlines the orchestration of multiple zones to preserve stability.

Reasoning zones enable safe, local handoffs to downstream reasoning engines without assuming global coherence. We adopt the reasoning-instance tuple $R=(P,E,f,g,\Pi)$ from \cite{nikooroo2025_reasoning_system} and define a zone-scoped instantiation.

We first define the internal structure of a reasoning instance restricted to a single $\theta$-zone.

\subsection{Zone-scoped reasoning}

Given a $\theta$-zone $Z\subseteq V_\theta$ (Sec.~\ref{sec:reasoning_zones}), define
\begin{gather}
R_Z \;=\; (P_Z, E_Z, f_Z, g_Z, \Pi_Z),
\end{gather}
where:
\begin{itemize}
    \item $P_Z := Z$ denotes the set of propositional claims restricted to $Z$,
    \item $E_Z := G[Z]$ is the subgraph of typed/directed edges induced by $Z$,
    \item $f_Z$ is an admissible inference or evaluation map (e.g., deductive closure or optimization objective) operating only on $(P_Z, E_Z)$,
    \item $g_Z$ optionally generates candidate conclusions within $Z$,
    \item $\Pi_Z$ encodes policies or principles that constrain or guide $f_Z$ and $g_Z$ (e.g., conservative vs.\ expansive inference).
\end{itemize}

To preserve modularity, each reasoning instance $R_Z$ is required to satisfy a logical isolation condition:
\begin{gather}
\big(p \vdash q \text{ under } R_Z \big)\ \Rightarrow\ \{p,q\} \subseteq Z,
\end{gather}
ensuring that conclusions do not leak outside the zone. Any intended cross-zone effect must be returned via explicit handback channels (defined below); no direct edits to beliefs or edges beyond $Z$ are permitted.

With these  instances specified, we now describe how their outputs are reconciled with the global belief graph.

\subsection{Handback channels to the belief graph}

Outputs of $R_Z$ are translated into candidate graph updates via two channels:

\begin{itemize}
\item \textbf{Credibility handback} $\Delta\Psi_Z$: proposals to adjust external priors on nodes in $Z$ (e.g., promote demoted premises). Implemented as a base-prior update $\mathbf{b}\leftarrow \mathrm{mix}(\mathbf{b},\Delta\Psi_Z)$ (convex mixing with stated weights) before re-running $T$ (Sec.~\ref{subsec:prop}).
\item \textbf{Structural handback} $\Delta E_Z$: proposals to add/retune edges within $Z$ (e.g., add a \texttt{support} edge for a proved implication, or weaken a contradicted link). Implemented as edits to $(\supp,\contr)$ on $E_Z$, followed by renormalization and recomputation of $\Phi^\star$. If $\Delta E_Z$ increases contradiction, the safety check from Sec.~\ref{sec:shock_updates} (contractivity  backtracking) applies.
\end{itemize}

\noindent Both handbacks are guarded by damping/line-search so the post-update operator remains contractive. After applying handbacks, refresh the atlas with Algorithm \ref{alg: ATLAS-UPDATE} (Sec.~\ref{sec:zone_governance}).

Finally, we describe how to run reasoning across the entire atlas, resolve conflicts between overlapping zones, and maintain safety and determinism during updates.

\subsection{Cross-zone orchestration}

Let $\mathcal{A}_\theta=\{Z_1,\ldots,Z_m\}$ be the reported atlas at threshold $\theta$. Reasoning is run independently within each zone, and the resulting edits are then merged deterministically:

\begin{center}
\begin{minipage}{0.90\linewidth} 
\begin{algorithm}[H]
\small
\caption{RUN-REASONING$\,(\mathcal{A}_\theta)$}
\begin{algorithmic}[1]
\For{$Z\in \mathcal{A}_\theta$}
   \State $(\Delta\Psi_Z,\Delta E_Z)\leftarrow R_Z(P_Z,E_Z,\Pi_Z)$
\EndFor
\State // conflict resolution across overlapping zones
\State Collect all proposed node prior changes and edge edits keyed by target
\For{each target $t$ with multiple proposals}
   \State keep the proposal from the zone with larger $S(Z)$ (Sec.~\ref{sec:zone_governance});
   \State ties: prefer larger $\sum_{v\in Z}\Phi^\star_v$, then smaller $\mathrm{cut}_{-}(Z)$, then lexicographic zone id
\EndFor
\State Apply the resulting $\Delta\Psi,\Delta E$ with damping/line-search to preserve contractivity
\State Recompute $\Phi^\star$ via $T$ and refresh $\mathcal{A}_\theta$ with Algorithm \ref{alg: ATLAS-UPDATE}
\end{algorithmic}\label{alg: RUN-REASONING}
\end{algorithm}
\end{minipage}
\end{center}

If a conclusion requires premises from $Z_i \cup Z_j$ and $\widetilde{G}_\theta[Z_i \cup Z_j]$ is balanced, the orchestrator may schedule a fused reasoning pass on $Z_i \cup Z_j$. Otherwise, the conclusion remains local. Fusion proposals follow the same handback and safety checks as ordinary runs.

Given fixed values of $(\theta,\tau)$, governance weights $(\lambda,\rho)$, and a chosen scoring rule, the aggregation and tie-break procedures yield deterministic and idempotent outcomes. Re-running Algorithm~\ref{alg: RUN-REASONING} on unchanged inputs produces the same atlas and graph state.
Under the isolation contract and the shock safety condition (Sec.~\ref{sec:shock_updates}), the algorithm alters $\Phi^\star$ outside $\mathcal{A}_\theta$ only through the propagation map $T$, which preserves contractivity and rules out oscillations.

%

\section{Credibility vs.\ confidence}
\label{sec:cred_vs_conf}

A defining feature of our framework is the explicit separation between \emph{credibility} and \emph{confidence}. They play different roles in the pipeline and must not be conflated.

\subsection{Credibility: source-oriented input trust}

Credibility, $\Psi(v)\in[0,1]$, is assigned at introduction based on provenance (e.g., authority, history, external knowledge-base priors). It is not propagated as a signal through the graph. Instead, credibility influences the \emph{base prior} $\mathbf{b}$ used by the confidence operator $T$ (Sec.~\ref{subsec:prop}):
\begin{gather}
\mathbf{b} = \lambda\,\frac{\Psi}{\|\Psi\|_\infty} + (1-\lambda)\,\mathbf{b}_0,\;\;\; \lambda\in[0,1].
\end{gather}
Modifying $\Psi$ perturbs $\mathbf{b}$, after which confidence is recomputed via $T$. Since $\Psi$ is normalized by $\|\Psi\|_\infty$, scaling all credibility scores by a positive constant has no effect.

We also consider a structure-derived prior, where $b_i = \sum_j \widehat{A}^+_{ij}$ represents the outgoing normalized support mass. All reported experiments specify which prior is used—credibility-based or structure-based.

\subsection{Confidence: structure-induced epistemic stability}

Confidence, $\Phi^\star(v)\in[0,1]$, is defined as the \emph{fixed point} of the signed, damped propagation map (Sec.~\ref{subsec:prop}). It reflects reinforcement or undermining due to the structure and weights of support and contradiction edges. By design, $\Phi^\star$ depends only on $(\widehat{A}^+, \widehat{A}^-)$ and the base prior $\mathbf{b}$; there is no separate credibility propagation mechanism.

The propagation dynamics are monotonic in $\mathbf{b}$: if $\mathbf{b}_1 \le \mathbf{b}_2$ elementwise (e.g., reflecting a credibility increase), then $\Phi^\star(\mathbf{b}_1) \le \Phi^\star(\mathbf{b}_2)$ (see Sec.~\ref{subsec:prop}). This property is useful for implementation audits and sanity checks.

\subsection{Interaction and divergence}

Credibility and confidence can align or diverge depending on the structural context. For example, a node with high credibility $\Psi(v)$ may still receive strong contradiction and therefore settle at low confidence $\Phi^\star(v)$. Conversely, a node with initially low $\Psi(v)$ may become well-supported within a balanced subgraph, leading to a high $\Phi^\star(v)$.

To avoid distortion, we impose a strict no-leakage policy: credibility scores are not propagated across the graph. Their only role is in forming the prior $\mathbf{b}$; the edge structure itself must reflect evidential relationships, not source trust. Encoding credibility into edges would constitute a form of double-counting, which we disallow.

\subsection{Implications for reasoning and updates}

Zone qualification depends on the inferred confidence values $\Phi^\star$, not the input credibility scores $\Psi$. Thresholding and balance criteria (see Sec.~\ref{sec:reasoning_zones}) are applied to $\Phi^\star$ when constructing reasoning zones.

When the system is externally updated, the two input channels are treated separately. Human or external judgments affect $\Psi$ and thus modify the prior $\mathbf{b}$. Structural changes such as shocks or inferred contradictions adjust the edge weights $(\supp, \contr)$. Both update types are guarded to preserve contractivity, using damping or line search as needed.

Reasoning inside a zone produces controlled outputs in one of two forms: (i) changes to $\Psi$, denoted $\Delta\Psi_Z$, or (ii) changes to the local edge structure, $\Delta E_Z$. These are returned through formal handback channels (Sec.~\ref{sec:interfaces}), after which the global confidence propagation is re-run.

\subsection{Internal checks and structural variants}

We conduct a set of internal tests that clarify system behavior under variant inputs and parameter regimes:

First, to test sensitivity to prior design, we perform a prior swap: the structure-based prior $b_i = \sum_j \widehat{A}^+_{ij}$ replaces the default, and changes to key metrics such as zone recovery, confidence thresholds $t^\star$, or post-shock stability are reported.

Second, to verify normalization behavior, we scale the credibility scores by a constant factor $c \in \{0.5, 2.0\}$. Since $\|\Psi\|_\infty$ normalization is applied, this scaling should not affect outcomes beyond numerical precision.

Finally, to detect any unintended leakage from credibility to edge structure, we set $\lambda=0$ (removing credibility input entirely) and confirm that edge weights remain unchanged. Any change would indicate an implementation flaw where $\Psi$ has inappropriately influenced $(\supp, \contr)$.

%

%

\section{Belief Update and System Evolution}
\label{sec:update_evolution}

Belief systems evolve in response to new inputs or shifts in internal structure. Updates affect the tuple
\[
\mathcal{B} = (V, E, \mathcal{T}, \Psi, \Phi^\star),
\]
through two levers: 1) \emph{Credibility inputs} (\(\Psi\)), which define the prior vector \(\mathbf{b}\), and 2) \emph{Graph structure} (edge types and weights), which influence propagation.
Each update triggers a recomputation of the fixed point \(\Phi^\star\) via the damped map \(T\), followed by a refresh of the zone atlas.

\subsection{Primitive operations}
The model supports three types of primitive updates: expansion, contraction, and revision. These operations affect the structure and/or the priors of the belief graph and trigger downstream updates in the propagation system, as follows:

\noindent In an \emph{expansion}, a new node \(v\) is added along with its credibility score \(\Psi(v)\) and typed incident edges. This perturbs both the base prior \(\mathbf{b}\) and the adjacency matrices \((\supp, \contr)\). Since row normalization is applied after insertion, rows that remain all-zero are left intact, and only the affected rows are renormalized.

\noindent A \emph{contraction} removes a node or a subset of its edges. This reduces outgoing or incoming mass and can alter the connected components of the graph. Normalization is again local, limited to the edited rows.

\noindent In a \emph{revision}, the credibility of an existing node \(\Psi(v)\) is adjusted, which updates the base prior \(\mathbf{b}\), and/or the edge types or weights are modified. Retyping follows the signed mapping convention from Sec.~\ref{sec:belief_graphs}, preserving nonnegativity by construction.

\subsection{Confidence recomputation and stopping}
Following any edit \(\Delta\), the updated confidence values \(\Phi^\star\) are obtained by recomputing the fixed point of the propagation map:
\begin{gather}
\Phi^\star \;=\; \lim_{t\to\infty} T^{(t)}(\mathbf{x}^{(0)}),\nonumber\\
T(\mathbf{x}) = \sigma\!\Big((1-\alpha)\mathbf{b} + \alpha(\widehat{\mathbf{A}}^{+} - \eta\widehat{\mathbf{A}}^{-})\mathbf{x}\Big),
\end{gather}
where contractivity is ensured by verifying the condition \(\alpha\|\widehat{\mathbf{A}}^{+}-\eta\widehat{\mathbf{A}}^{-}\|_2<1\) (see Theorem~\ref{thm:contraction}). Iteration stops when the change between successive steps falls below a tolerance, \(\|\mathbf{x}^{(t+1)}-\mathbf{x}^{(t)}\|_\infty\le\varepsilon\), or when a maximum iteration count \(t\ge T_{\max}\) is reached. If contractivity is violated, a damping or line search routine (Sec.~\ref{sec:shock_updates}) is applied to reduce the magnitude of \(\Delta\) or adjust \(\alpha\) until the condition is restored.

Because \(T\) is a contraction under the stated condition, the fixed point \(\Phi^\star\) is guaranteed to exist and be unique, regardless of initialization. In practice, we re-initialize from the pre-edit fixed point to accelerate convergence.

\subsection{Zone reconfiguration and atlas maintenance}
After \(\Phi^\star\) updates,  maximal \(\theta\)-zones are extracted via Algorithm \ref{alg: EXTRACT-ZONES} (Sec.~\ref{sec:reasoning_zones}) and  Algorithm \ref{alg: ATLAS-UPDATE} with Policy~\ref{policy:atlas} (Sec.~\ref{sec:zone_governance}) is applied.
Zones may \emph{emerge}, \emph{morph}, or \emph{collapse} as \(\Phi^\star\) shifts.
We use hysteresis (Sec.~\ref{sec:zone_governance}) to reduce churn, and the local-refresh variant (re-extract within \(h\) hops of nodes whose confidence crossed \(\theta\)) to limit work to affected regions.
\subsection{Adaptive Handling of Contradiction}

Contradictions are introduced through shock updates (Sec.~\ref{sec:shock_updates}), which modify the \((\supp, \contr)\) terms on affected nodes, then recheck contractivity before recomputing \(\Phi^\star\). A backtracking safeguard ensures that updates remain stable and prevents oscillatory behavior. Once shock effects are absorbed, several zone-level strategies can govern how the system resolves or tolerates conflicting evidence. One option is \emph{pluralism}, allowing multiple overlapping or competing zones to coexist, with final inclusion controlled by a Jaccard threshold \(\tau\) and scored via confidence-weighted measures \(S(Z)\). Alternatively, \emph{suppression} may downweight outgoing influence from low-confidence nodes or zones through soft multiplicative decay, as long as the contractivity condition holds. Finally, targeted \emph{prior steering} can adjust the credibility input \(\Psi\) for adjudicated nodes, shifting the prior \(\mathbf{b}\) without inducing circular trust propagation.

\subsection{Deterministic update routine}
 Algorithm \ref{alg: UPDATE-AND-REFRESH} executes a full update cycle after an edit \(\Delta\), ensuring stable propagation and consistent extraction of reasoning zones. The procedure guarantees determinism: given fixed hyperparameters and tie-breaking rules, rerunning the routine on identical inputs yields the same fixed point \(\Phi^\star\) and the same zone atlas \(\mathcal{A}_\theta\).

\begin{center}
\begin{minipage}{0.92\linewidth}
\begin{algorithm}[H]
\small
\caption{UPDATE-AND-REFRESH\\$(\Delta;\theta,\tau,\alpha,\eta)$}
\begin{algorithmic}[1]
\State Apply \(\Delta\) to \(\Psi\) and/or \((\supp,\contr)\); renormalize affected rows to get \(\widehat{A}^{+},\widehat{A}^{-}\)
\While{\(\alpha\|\widehat{A}^{+}-\eta\widehat{A}^{-}\|_2 \ge 1\)}
   \State damp \(\Delta\) (or reduce \(\alpha\)) via backtracking; renormalize
\EndWhile
\State \(\Phi^\star \leftarrow\) iterate \(T\) from previous \(\Phi^\star\) until \(\|\Delta\|_\infty\le\varepsilon\) or \(t\ge T_{\max}\)
\State \(\mathcal{A}_\theta \leftarrow \textsc{EXTRACT-ZONES}(\Phi^\star,\theta)\); \(\mathcal{A}_\theta \leftarrow \textsc{ATLAS-UPDATE}(\tau)\)
\State \textbf{return} \(\Phi^\star,\mathcal{A}_\theta\)
\end{algorithmic}\label{alg: UPDATE-AND-REFRESH}
\end{algorithm}
\end{minipage}
\end{center}

\section{Evaluation Plan}
\label{sec:evaluation}

We evaluate (i) propagation well-posedness and convergence, (ii) zone extraction quality, and (iii) atlas governance under overlap, (iv) robustness to contradiction shocks. These experiments serve as a first exploration of the framework’s mechanics rather than a comprehensive benchmark, establishing a baseline for future extensions.

\subsection{Graph generators}
Unless stated otherwise, $n\in\{400,800,1600\}$, target sparsity $p\in\{0.01,0.02,0.04\}$, edge weights are i.i.d.\ $\mathrm{LogNormal}(\mu{=}0,\sigma{=}0.5)$, and node credibilities follow a Pareto tail ($\alpha{=}2.5$) rescaled to $[0,1]$.

\textbf{G1: Signed ER with acyclic bias}:
Start from $G(n,p)$ and orient edges from lower to higher node id to induce few cycles (DAG-like backbone); flip $10\%$ of edges to random directions to allow short cycles. Assign signs i.i.d.\ with $P(+){=}0.7$, $P(-){=}0.3$; weights as above.

\textbf{G2: Planted balanced zones}:
Sample $k\in\{3,5\}$ disjoint blocks with sizes uniform in $[0.05n,0.12n]$. Inside each block, draw edges with sign $+$ w.p.\ $0.9$ and $-$ w.p.\ $0.1$ but \emph{repair} any negative odd cycle by flipping its lightest edge to $+$ (guaranteeing balance). Between blocks/background: $P(+){=}0.6$, $P(-){=}0.4$. Increase node credibilities inside blocks by $+0.15$ (clipped to $1$) to emulate higher-quality sources. Ground-truth zones $\{Z_i^\star\}$ are the planted blocks.

\textbf{G3: Stress graphs near the contraction boundary}:
Start from G1 and (i) add $h=\lceil 0.01n\rceil$ hub nodes with degree $\approx 0.2n$, and (ii) sprinkle $0.02|E|$ disjoint negative $3$-cycles along block boundaries (if present) to probe extraction repair and shock safety.

\subsection{Protocols and metrics}

Unless stated otherwise, we use the following default hyperparameters: confidence threshold $\theta\in\{0.6,0.7,0.8\}$; Jaccard de-duplication threshold $\tau\in\{0.2,0.3,0.4\}$ (default $0.30$); and atlas truncation to top-$k$ with $k\in\{3,5\}$.

We evaluate our framework across four protocol types, denoted P1–P4, each targeting a specific functional property of the system.

\vspace{0.5em}
\noindent\textbf{Propagation validity (P1)}:  
We explore a grid of parameters $\alpha\in\{0.2,0.4,0.6,0.8\}$ and $\eta\in\{0,0.5,1\}$, using a structure-based prior with $b_i=\sum_j\widehat{A}^+_{ij}$. For each setting, we report:
\begin{itemize}
\item the spectral factor $r=\alpha\|\widehat{A}^+-\eta\widehat{A}^-\|_2$, estimated via power iteration (3 random starts, 200 steps, $\ell_2$ norm);
\item convergence iterations $t^\star = \min\{t : \|x^{(t)} - x^{(t-1)}\|_\infty \le \varepsilon\}$ with $\varepsilon=10^{-6}$ and cap $T_{\max}=2000$.
\end{itemize}

\vspace{0.5em}
\noindent\textbf{Zone extraction quality (P2, G2).}  
We compute the recovered atlas $\widehat{\mathcal{A}}_\theta$ and evaluate best-match accuracy to planted zones using the Hungarian algorithm on $1{-}\mathrm{Jaccard}$ distances. The core metrics are:
\begin{gather}
\mathrm{Precision} = \frac{1}{|\widehat{\mathcal{A}}_\theta|} \sum_{Z\in\widehat{\mathcal{A}}_\theta} \max_i \frac{|Z \cap Z_i^\star|}{|Z|},\\
\mathrm{Recall} = \frac{1}{|\{Z_i^\star\}|} \sum_i \max_{Z} \frac{|Z \cap Z_i^\star|}{|Z_i^\star|}, \quad F_1 = \frac{2PR}{P+R}.\nonumber
\end{gather}
Additional diagnostics include: (i) number of negative-parity cycles inside zones (should be zero); and (ii) zone confidence statistics.

\vspace{0.5em}
\noindent\textbf{Governance under overlap (P3).}  
We vary the de-duplication threshold $\tau\in\{0.2,0.4,0.6\}$ to assess atlas stability and coverage. Evaluation covers:
\begin{itemize}
\item atlas size $|\mathcal{A}_\theta|$, coverage $\frac{|\cup_{Z\in\mathcal{A}_\theta}Z|}{|V_\theta|}$, and mean inter-zone Jaccard;
\item zone stability $S_J = \mathrm{mean}_Z\left[\max_{Z'\in\mathcal{A}_\theta^{\text{ref}}} J(Z,Z')\right]$ when sweeping $\tau$, with reference $\tau{=}0.30$.
\end{itemize}

\vspace{0.5em}
\noindent\textbf{Shock robustness (P4, G1/G3).}
We test resilience under batch shocks, distributing mass $m\in\{0.1,0.2,0.4\}$ across $|U|=\lfloor 0.05n \rfloor$ uniformly sampled nodes. Contractivity is maintained via backtracking. Metrics include:
\begin{itemize}
\item \textbf{Zone stability} $S_J$ pre- and post-shock, relative to a fixed reference atlas;
\item \textbf{False-collapse rate} (G2): fraction of planted zones $Z_i^\star$ that lose all matching zones with Jaccard $\ge \tau{=}0.3$ post-shock, despite $70\%$ node-level retention ($x^\star_u \ge \theta$).
\end{itemize}

Each point averages over $30$ seeds; bars show $95\%$ CIs (normal approximation on seed means). We fix $(\varepsilon{=}10^{-6}, T_{\max}{=}2000)$, power-iteration settings (3 starts, 200 steps), and defaults $(\theta{=}0.7,\tau{=}0.30,k{=}3,\alpha{=}0.6,\eta{=}0.5)$. We release:
\begin{itemize}
\item scripts for G1–G3, the extractor, governance, and shocks;
\item config files (YAML) for all grids and seeds; CSV schemas per protocol (including $r$, $t^\star$, wall-clock, $|\mathcal{A}_\theta|$, coverage, mean Jaccard, $S_J$, false-collapse rate);
\end{itemize}

%

\section{Evaluation}
\label{sec:evaluation}

We evaluate four aspects of the framework using {synthetic, ground-truthed graphs}: 
(P1) well-posed propagation, 
(P2) zone extraction quality, 
(P3) governance stability under small perturbations, 
and (P4) robustness to exogenous shocks. 
Unless otherwise stated we use \(n=2000\) nodes and \(30\) seeds.

The goal of this evaluation is diagnostic rather than competitive: it demonstrates the framework’s stability and interpretability on controlled synthetic settings, leaving broader and real‑world assessments to forthcoming work.

\textit{Signed graph}:
We split the row-weighted adjacency into positive and negative parts,
\(\widehat{\mathbf A}^{+}\!\ge 0\) and \(\widehat{\mathbf A}^{-}\!\ge 0\).
For each node \(i\), outgoing positive (resp.\ negative) weights are normalized so their row sums satisfy 
\(\|\widehat{\mathbf A}^{+}_{i:}\|_{1}\le 1\) and \(\|\widehat{\mathbf A}^{-}_{i:}\|_{1}\le 1\); 
rows with no outgoing edges remain zero.

\textit{Propagation map}:
Let \(M(\eta) = \widehat{\mathbf A}^{+} - \eta\,\widehat{\mathbf A}^{-}\).
Given a bias vector \(\mathbf b \in [0,1]^n\) and damping \(\alpha \in (0,1)\), iterate
\[
\mathbf x_{t+1} = (1-\alpha)\,\mathbf b + \alpha\,M(\eta)\,\mathbf x_t,\qquad 
\mathbf x_0 = \mathbf b.
\]
Stop at the first \(t^\star\) with \(\|\mathbf x_{t+1}-\mathbf x_t\|_\infty\le 10^{-6}\) (cap \(t\le 2000\)). 
\(t^\star\) and \(\Phi^\star=\mathbf x_{t^\star}\) are reported.

\textit{Contraction check}:
Define \(r=\alpha\,\|M(\eta)\|_2\).
Estimate \(\|\,\cdot\,\|_2\) by power iteration (3 random starts, 200 steps, tolerance \(10^{-6}\)); 
if \(r<1\) the map is a contraction.

\textit{Zone extraction and atlas update}:
Given \(\Phi^\star\) and threshold \(\theta\), let \(V_\theta=\{i:\Phi^\star_i\ge \theta\}\).
Form the undirected, signed \emph{dominant-sign} projection on \(G[V_\theta]\) 
by aggregating both directions between each pair and assigning the sign with larger total weight. 
Candidates are the connected components of this projection; 
each candidate is tested for signed-cycle \emph{balance} via signed 2-coloring. 

\noindent If unbalanced, apply a greedy repair (remove the minimum-confidence vertex on the certificate cycle, recurse). 
Atlas de-duplication uses Jaccard \(J(Z,Z')=\frac{|Z\cap Z'|}{|Z\cup Z'|}\) 
with overlap threshold \(\tau=0.30\) and keeps at most \(k\) zones (here \(k=3\)) by descending \(\sum_{i\in Z}\Phi^\star_i\).

\textit{Ground-truth matching and metrics}:
For a reported family \(\widehat{\mathcal A}\) and truth \(\{Z_j^\star\}_{j=1}^k\):
\begin{align*}
\text{Prec} &= \frac{1}{|\widehat{\mathcal A}|} \sum_{Z\in\widehat{\mathcal A}} 
\max_j \frac{|Z\cap Z^{\star}_j|}{|Z|}, \\
\text{Rec}  &= \frac{1}{k} \sum_j \max_{Z\in\widehat{\mathcal A}} 
\frac{|Z\cap Z^{\star}_j|}{|Z^{\star}_j|}, \\
F_1 &= \frac{2\,\text{Prec}\cdot\text{Rec}}{\text{Prec}+\text{Rec}}.
\end{align*}

Node-level scores use \(\widehat{U}=\bigcup_{Z\in\widehat{\mathcal A}} Z\) and \(U^\star=\bigcup_j Z_j^\star\).

Churn under change is 
\[
\chi=1-\frac{1}{|\mathcal A_{\text{pre}}|}\sum_{Z\in\mathcal A_{\text{pre}}}\max_{Z'}J(Z,Z').
\]

Shock stability is 
\[
S_J=\frac{1}{|\mathcal A_{\text{pre}}|}\sum_{Z\in\mathcal A_{\text{pre}}}\max_{Z'}J(Z,Z').
\]

Across seeds we report mean \(\pm\)95\% CI, i.e., \(1.96\cdot \text{sd}/\sqrt{5}\).

\textit{Graph families}: All graphs are directed; raw weights are i.i.d.\ \(\mathrm{TruncNormal}(\mu=1,\sigma=0.2; w>0)\) 
then row-normalized \emph{by sign} as above. Randomness is controlled by the stated seed.

\begin{description}
\item[G1: Signed random (P1).]
Expected out-degree \(d=8\). Each node chooses \(d\) distinct out-neighbors uniformly; 
an edge is negative with probability \(\rho_-=0.3\), positive otherwise. 
\item[G2: Planted balanced zones (PBZ; P2--P3).]
Choose \(k=3\) disjoint zones of size \(s=\max(120,n/10)\) (so \(s=200\) when \(n=2000\)). 
Inside each zone: add positive edges with \(p_{\text{in}}=0.22\); 
negative edges inside are forbidden (balanced interior).
Outside zones: add positives with \(p_{\text{out,pos}}=0.01\) and negatives with \(p_{\text{out,neg}}=0.01\).
\item[G3: Stress graphs (P4).]
Start from G1 with \(d=8\), then embed \(C=\max(50,n/20)\) disjoint negative 3-cycles, then normalize by sign.
\end{description}

\textit{Bias vector}:
For all runs \(b_i\) equals the row-sum of \(\widehat{\mathbf A}^{+}\) in row \(i\), 
i.e., the node’s outgoing positive mass (structure-based prior).

\textbf{\textit{Baselines}}: Where applicable, the proposed approach is benchmarked against the following schemes to highlight the added value of signed propagation and balance constraints:
\begin{itemize}
    \item \textbf{UnsignCL}: Louvain clustering on the unsigned thresholded graph \( G_\theta \), treated as reasoning zones.
    \item \textbf{UnsignPRO}: Unsigned propagation (e.g., PageRank) followed by the same extraction pipeline.
\end{itemize}
 
\noindent Future versions will expand to richer signed-reasoning baselines as the architecture matures.

\subsection*{Tasks and results}
We now summarize key findings from each protocol (P1–P4), using the corresponding graph families and metrics outlined above.

\begin{figure}[t]
    \centering
    \includegraphics[width=1\linewidth]{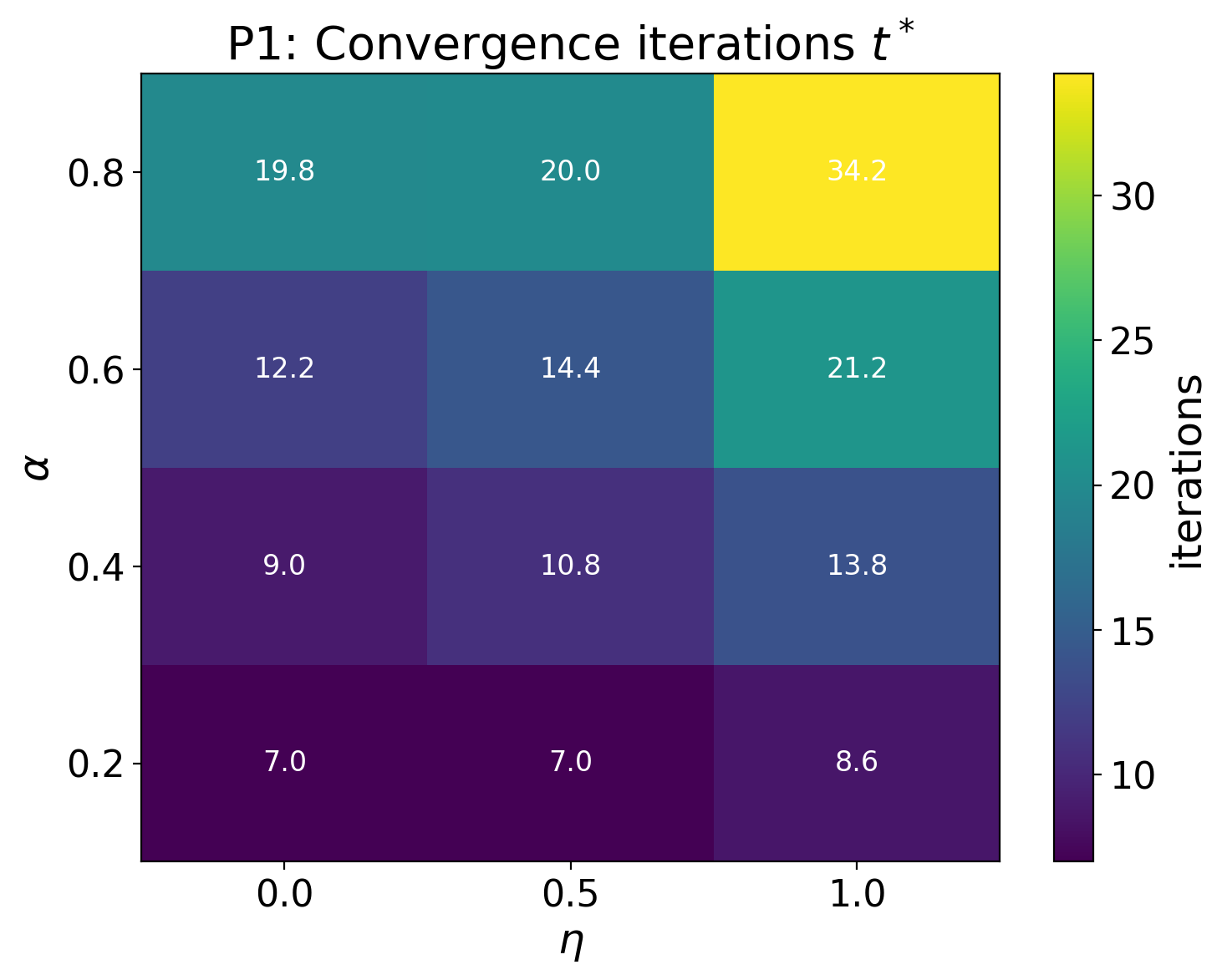}
    \caption{P1: iterations to tolerance $t^\star$ vs.\ $(\alpha,\eta)$.}
    \label{fig:p1}
\end{figure}

\textit{P1: Propagation validity}:
Graphs: G1. Grid: \(\alpha\in\{0.2,0.4,0.6,0.8\}\), \(\eta\in\{0,0.5,1\}\). For each seed and \((\alpha,\eta)\): compute \(r=\alpha\|M(\eta)\|_2\) (power iteration); run the fixed-point iterate from \(\mathbf x_0=\mathbf b\) to tolerance \(10^{-6}\) (cap 2000 iters); record \(t^\star\). Figure~\ref{fig:p1} shows that convergence slows near the contractivity boundary, especially at \((\alpha,\eta)=(0.8,1.0)\), but remains stable.

\begin{figure}[t]
    \centering
    \includegraphics[width=1\linewidth]{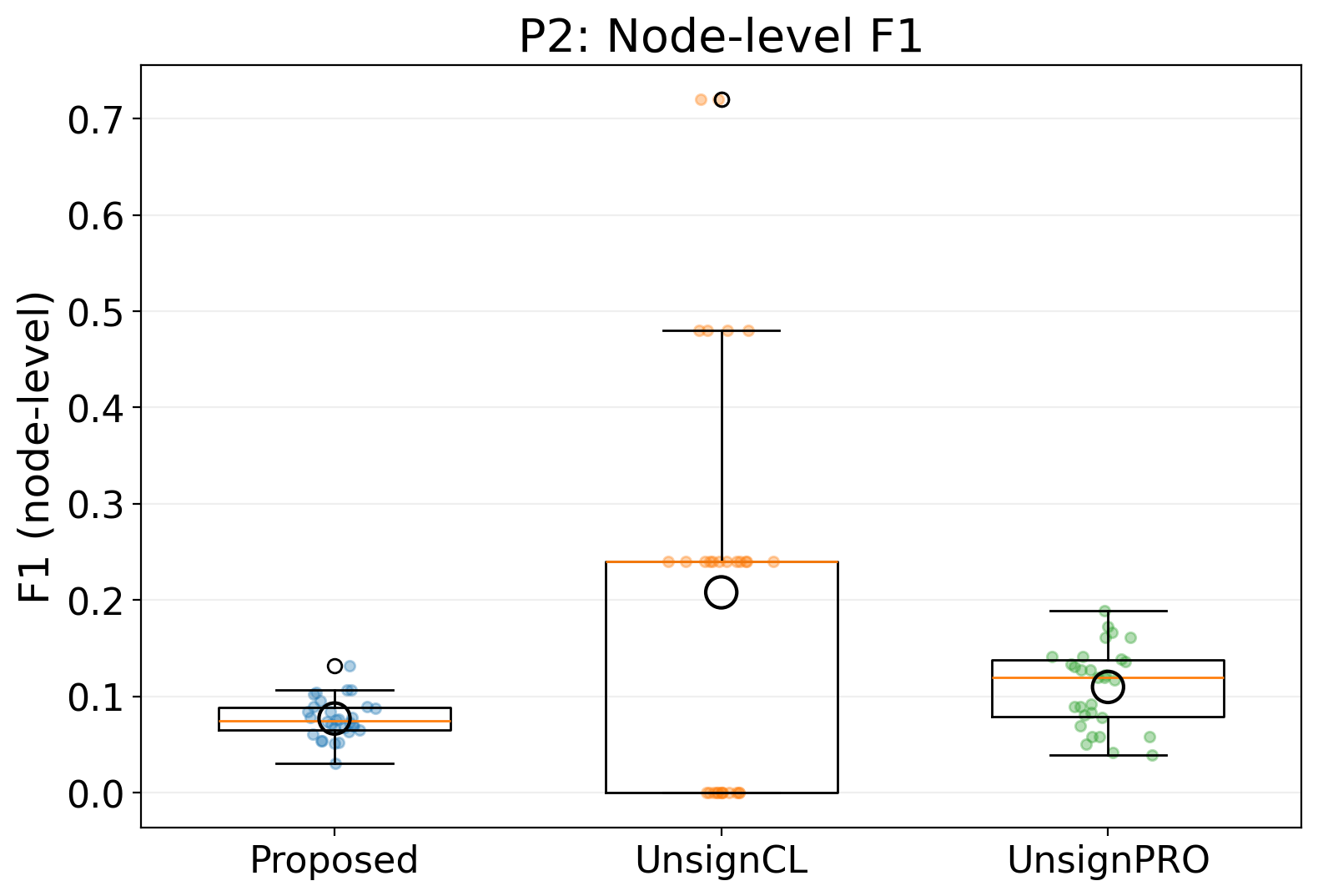}
    \caption{P2: node-level $F_1$ by method (mean $\pm$95\% CI).}
    \label{fig:p2node}
\end{figure}

\begin{figure}[t]
    \centering
    \includegraphics[width=1\linewidth]{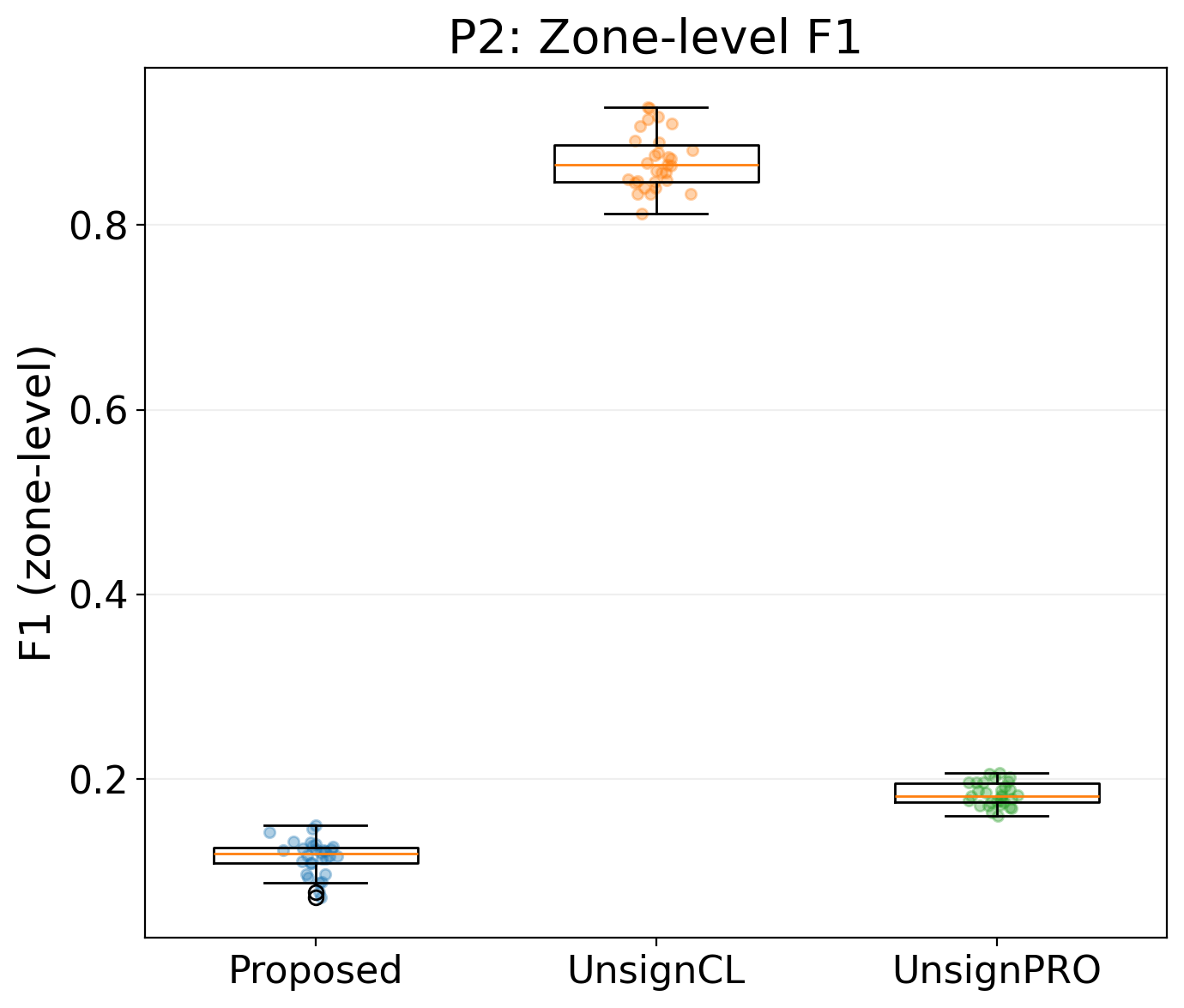}
    \caption{P2: zone-level $F_1$ by method (mean $\pm$95\% CI).}
    \label{fig:p2zone}
\end{figure}

\textit{P2: Zone extraction quality}:
Graphs: PBZ (G2); propagation \((\alpha,\eta)=(0.6,1.0)\). Threshold sweep: \(q\in\{0.30,\dots,0.90\}\) with \(\theta(q)=\mathrm{quantile}(\Phi^\star,q)\). Zones extracted, balance and repair tested,  atlas updated \((\tau=0.30)\), while keeping \(k=3\). We select \(q^\star\) that maximizes node-level \(F_1\). Figure~\ref{fig:p2node} shows node-level \(F_1\) by method; Figure~\ref{fig:p2zone} shows zone-level \(F_1\). All reported zones were balanced after repair. 

\noindent At the node level (Figure~\ref{fig:p2node}), {UnsignPRO} shows the highest central tendency (\(0.11{-}0.12\)), {The proposal} scores lower but is tight (stable across seeds), while {UnsignCL} is {unstable} (many near-zero runs with occasional high outliers). In Figure~\ref{fig:p2zone}, {UnsignCL} attains the highest zone-level \(F_1\) (\(0.85{-}0.90\) median), reflecting that unsigned community structure closely matches the planted (mostly-positive) interiors. {UnsignPRO} is substantially lower (\(0.17{-}0.20\)), and the {Proposed} method is lowest (\(0.10{-}0.13\)), consistent with stricter balance and confidence gating (conservative recall).  In short, the unsigned clustering baseline excels on these planted, largely positive interiors, whereas our method prioritizes balance-respecting, conservative inclusion; improving zone-level recall without sacrificing signed consistency is left for follow-up tuning.

\begin{figure}[t]
    \centering
    \includegraphics[width=1\linewidth]{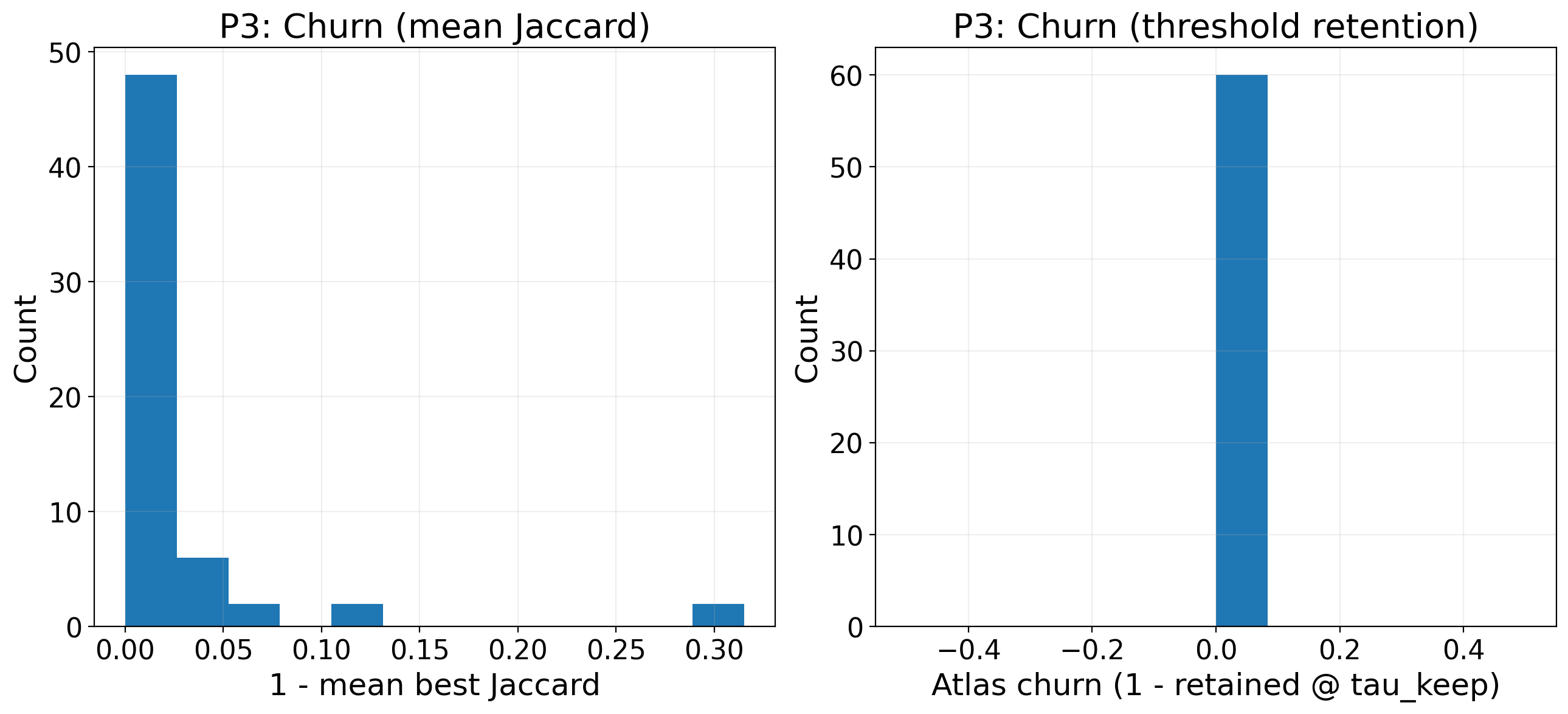}
    \caption{P3: governance churn under $5\%$ weight jitter. Left: mean Jaccard churn. Right: $\tau$-threshold churn.}
    \label{fig:p3}
\end{figure}

\textit{P3: Governance stability under jitter}:
Graphs: PBZ (G2); propagation \((\alpha,\eta)=(0.6,1.0)\); threshold at 0.75-quantile pre and post jitter. Apply weight jitter: \(w \mapsto w(1 + 0.05\mathcal{N}(0,1))\), clamp to \([0,\infty)\), no renormalization. Metric: atlas churn \(\chi\) via Jaccard change. Figure~\ref{fig:p3} shows churn histograms for both mean Jaccard and \(\tau\)-thresholded atlas.

\begin{figure}[t]
    \centering
    \includegraphics[width=1\linewidth]{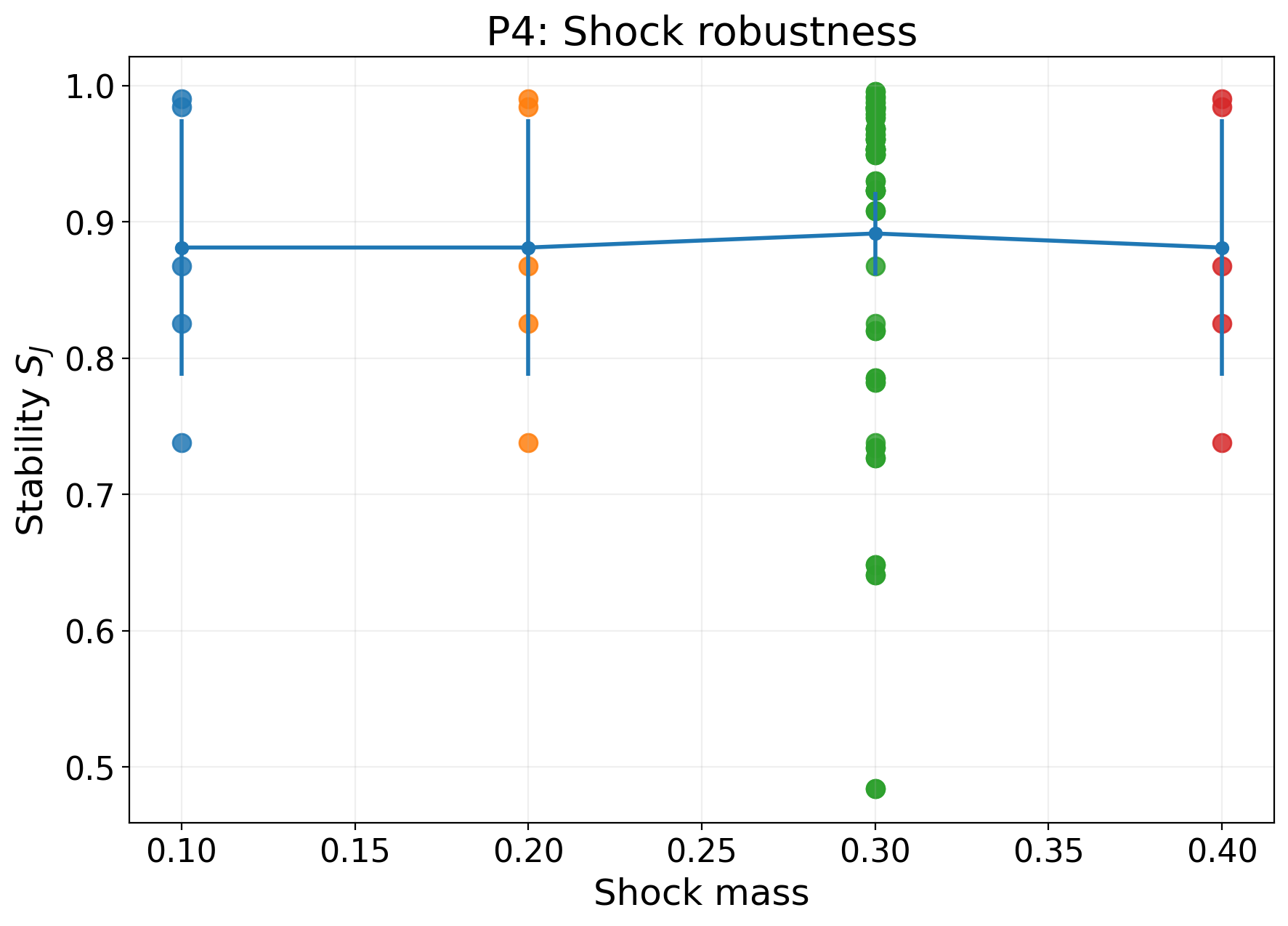}
    \caption{P4: robustness $S_J$ vs.\ shock mass $m$ (higher is better).}
    \label{fig:p4}
\end{figure}

\textit{P4: Shock robustness}:
Graphs: stress graphs (G3); propagation \((\alpha,\eta)=(0.6,1.0)\); threshold at 0.75-quantile. Apply shock \(s_u = m/40\) to 40 nodes per seed, for \(m\in\{0.1,0.2,0.3,0.4\}\); backtrack if contractivity is broken. Figure~\ref{fig:p4} shows zone-level stability \(S_J\) vs.\ shock mass. Robustness remains high across the board.

%

\section{Discussion and Future Directions}
\label{sec:discussion}

The framework provides a structural account of belief in which \emph{local} reasoning remains viable despite \emph{global} epistemic instability. By separating externally assigned \emph{credibility} from internally propagated \emph{confidence}, and by defining \emph{reasoning zones} (RZs) as confidence-qualified, signed-balance subgraphs, we obtain selective trust, adaptive inference, and tolerance to contradiction—properties that are difficult to realize in globally coherent models.
This work should be read as a first articulation of a broader research program—aimed at testing whether local coherence mechanisms can systematically substitute for global consistency in reasoning systems.

\subsection{Relation to Existing Models}

Our approach complements, rather than replaces, classical formalisms:
\begin{itemize}
    \item \textbf{Probabilistic belief (Bayesian/DS)}: These treat belief as a globally normalized scalar. Here, belief is \emph{structural}: support and contradiction edges shape emergent confidence through a contractive propagation map. Fragmentation and localized coherence arise naturally.
    \item \textbf{Belief revision (AGM)}: AGM typically seeks global consistency after updates. We instead tolerate persistent contradiction, routing inference through subgraphs that pass a \emph{signed-cycle balance} test; global inconsistency need not block local reasoning.
    \item \textbf{Argumentation frameworks}: Many frameworks define binary acceptance via attack/support structures. We retain signed edges but compute a \emph{continuous}, structure-induced confidence, and govern zone membership via Harary balance instead of extension semantics.
\end{itemize}

\subsection{Design Benefits}

Operationally, the framework supports selective inference and graceful degradation. 
Classical reasoning activates only within zones that meet confidence and balance criteria, while zones shrink or split under rising contradiction instead of triggering global collapse. 
Credibility remains local—it shapes only the base prior and never propagates—thus avoiding trust double-counting along edges. Failures localize to specific nodes or signed cycles, 
with atlas scores and overlaps making such trade-offs explicit. Finally, shock backtracking preserves contractivity, and atlas hysteresis ensures stable updates under small perturbations.

\subsection{Limitations and Threats to Validity}

The proposed framework carries several limitations. For balance testing, we aggregate both edge directions and retain the dominant sign. While this yields a fast and robust approximation, it ignores directionality, and stricter, direction-aware formulations may yield different behavior at higher computational cost. The results can also be sensitive to the choice of prior \( \mathbf{b} \); although we report the specific prior used and include ablations, the distinction between credibility-based and structure-based initialization may influence downstream inferences. Our evaluations rely on synthetic graphs with controlled structure, including planted balanced zones. This helps isolate mechanisms but may underrepresent real-world complexities such as heavy-tailed degree distributions, polarity-mixed communities, or annotation noise. Finally, the greedy repair step in zone extraction prioritizes speed over global optimality. While it produces high-quality zones in practice, the method does not guarantee optimal partitions in the worst case, making it a promising baseline for subsequent refinement and theoretical analysis.

\subsection{Technical Extensions}

Several extensions are natural. 
First, edges and priors can evolve over time through decay or reinforcement, allowing temporal dynamics and stability of zones to be studied under streaming updates. 
Second, the framework admits probabilistic fusion, where edge types or contradiction penalties are treated as random variables and the contractive map is coupled with probabilistic inference over such parameters. 
Third, layered graphs can be introduced to represent beliefs, meta-beliefs, and methodological constraints in separate strata, enabling reasoning both within and across layers under structured shocks. 
Finally, in multi-agent settings, distinct atlases may be merged or aligned across agents, supporting negotiation, conflict mediation, and trust calibration at the zone level.

\subsection{Open Epistemic Questions}
Several conceptual questions remain open: 
\begin{itemize}
    \item How do multiple high-confidence zones compete or cooperate under ambiguous evidence?
    \item Which graph features (e.g., signed assortativity, cut structure) predict zone stability or inferential yield? 
    \item What minimal signed structures suffice for coherent local reasoning to emerge under noise and contradiction?
\end{itemize}
These questions point toward deeper theoretical foundations and broader applicability in complex belief settings, and will guide the next iteration of this framework.


%

\section{Conclusion}
\label{sec:conclusion}

We presented a first graph-theoretic account of belief that cleanly separates \emph{credibility} (external, source-oriented priors) from \emph{confidence} (internal, structure-induced valuations). Confidence is computed by a normalized, damped propagation scheme with a stated contractivity condition, ensuring a unique, stable solution.

A central contribution is the notion of \emph{reasoning zones}: confidence-thresholded, signed-balance subgraphs extracted in near-linear time (dominant-sign projection, signed 2-coloring, and greedy repair) and curated into an \emph{atlas} with explicit overlap governance. This enables classical inference to operate safely and locally even in the presence of global contradiction. Furthermore, \emph{shock updates} was introduced that retune edges under a contractivity-preserving backtracking rule, yielding localized reconfiguration (shrink, split, collapse) rather than system-wide failure.

Through a worked example and a synthetic evaluation suite, we demonstrated well-posed propagation, accurate zone recovery, governance stability, and robustness to exogenous shocks. The framework is modular: it can hand off zones to downstream reasoning engines, accept handbacks as prior or structural edits, and maintain stability throughout via the same contractive operator.

Looking ahead, key directions include temporal dynamics, probabilistic fusion, layered belief representations, and multi-agent orchestration. A major next step is to connect reasoning zones to downstream inference tasks, enabling zone-scoped decision-making. Beyond engineering, the model poses epistemic questions about when and why coherent reasoning can emerge from contradictory substrates. We hope this serves as both a practical starting point and a conceptual foundation for future studies of selectively coherent reasoning in complex environments.

%


\printbibliography


\end{document}